# Automatic Arabic Dialect Identification Systems for Written Texts: A Survey

A Preprint

Maha J. Althobaiti
Department of Computer Science
Taif University
Taif, Saudi Arabia
maha.j@tu.edu.sa

September 22, 2020

## Abstract

Arabic dialect identification is a specific task of natural language processing, aiming to automatically predict the Arabic dialect of a given text. Arabic dialect identification is the first step in various natural language processing applications such as machine translation, multilingual text-to-speech synthesis, and cross-language text generation. Therefore, in the last decade, interest has increased in addressing the problem of Arabic dialect identification. In this paper, we present a comprehensive survey of Arabic dialect identification research in written texts. We first define the problem and its challenges. Then, the survey extensively discusses in a critical manner many aspects related to Arabic dialect identification task. So, we review the traditional machine learning methods, deep learning architectures, and complex learning approaches to Arabic dialect identification. We also detail the features and techniques for feature representations used to train the proposed systems. Moreover, we illustrate the taxonomy of Arabic dialects studied in the literature, the various levels of text processing at which Arabic dialect identification are conducted (e.g., token, sentence, and document level), as well as the available annotated resources, including evaluation benchmark corpora. Open challenges and issues are discussed at the end of the survey.

## 1 Introduction

The current era of intelligent language systems that perform many functions (e.g., machine translation, social media analysis, cyber security, and marketing) creates the necessity to computationally handle texts of different domains (e.g., news, blogs, Twitter messages, and customer reviews) and topics (e.g., economy, politics, sports, and science). Arabic text processing is one of the challenges that faces the researchers and developers of computational linguistics due to many factors. The first of these factors is that Arabic language generally refers to a collection of varieties with morphological, syntactic, lexical, and phonological differences [1]. These varieties include a standardized form, Modern Standard Arabic (MSA), and many non-standardized regional dialects. The MSA is mostly written and not spoken. That is, MSA is usually used in written texts, but is also used in formal events, sermons, education, political speeches, and news broadcasts [2, 3]. On the other hand, regional Arabic dialects are mainly spoken, not written, as they are utilized for informal daily communication, and also in radio and TV soap operas.

The regional dialects started to appear in a text form in the new millennium with the rise of Web 2.0, which allowed websites to have interactive contents generated by users (e.g., social media posts, blogs, emails, discussion forums). The online Arabic texts are less controlled, more speech-like, and usually written in an informal manner using colloquial dialects [4]. They are usually inconsistent, since Arabic dialects lack orthographic standards. Moreover, the Linguistic Code Switching (LCS) phenomenon appears in online Arabic content. That is, the writer of online texts sometimes switches between MSA and at least one Arabic dialect within the same utterance [5]. This makes processing the Arabic online texts computationally a challenging issue that should be addressed when building models for different Natural Language Processing (NLP) tasks. Moreover, many studies [6, 7, 8] have reported that tools built specifically for MSA resulted in significantly lower performance when applied to texts of Arabic dialects due to the significant linguistic differences between MSA and dialects. Therefore, more attention has recently been given to computational approaches to processing the texts written in Arabic dialects [9, 10, 7, 8].

Arabic Dialect Identification (ADI) in written texts is an active NLP task aiming to automatically identify the Arabic dialect of given texts. The availability of accurate Arabic dialect identification models can be of great benefit to many Arabic NLP tasks, such as statistical machine translation, natural language generation, and building dialect-to-dialect lexicons. The problem of ADI in written texts received a great deal of attention from researchers in the last decade, whereby computational approaches and system architectures have been developed to enhance ADI [11, 6, 12, 13, 14, 15], as shown in Figure 1. Therefore, there is a need to shed light on the currently functioning ADI and to direct new ADI research efforts to close the gaps. The purpose of this paper is to provide an informative survey of studies on Arabic Dialect Identification including available lexical resources, common benchmarks for evaluation, used features, adopted learning methods, implemented system architectures, and the considered Arabic dialects. The paper also discusses the current and potential applications and tools of ADI without forgetting outstanding issues and challenges, as well as the future of the ADI in written texts.

This paper presents a survey of ADI research in written texts and does not mention any work that focuses on speech alone. However, our paper references the studies conducted on speech transcripts and that utilize acoustic features in addition to lexical features. The remainder of the paper is organized as follows: Section 2 covers the Arabic dialects including their categories, characteristics, and how they differ from each other and from MSA in terms of lexicons, syntax, morphology and phonology. Section 3 presents the Arabic dialect identification task by defining the problem of ADI in written texts and then discussing the challenges and issues at hand. Section 4 covers the available Arabic dialect identification corpora required to build ADI models, as well as the common benchmarks used in the literature to evaluate the models. Section 5 presents the evaluation metrics utilized in the literature, shared tasks and ADI evaluation campaigns. Section 6 discusses the ADI studies, covering several aspects such as adopted features, utilized learning methods, components of the ADI systems, and the considered Arabic dialects. The section also incorporates qualitative comparison between existing works using the common benchmarks. Lastly, open issues and the future of Arabic dialect identification are presented in Section 7.

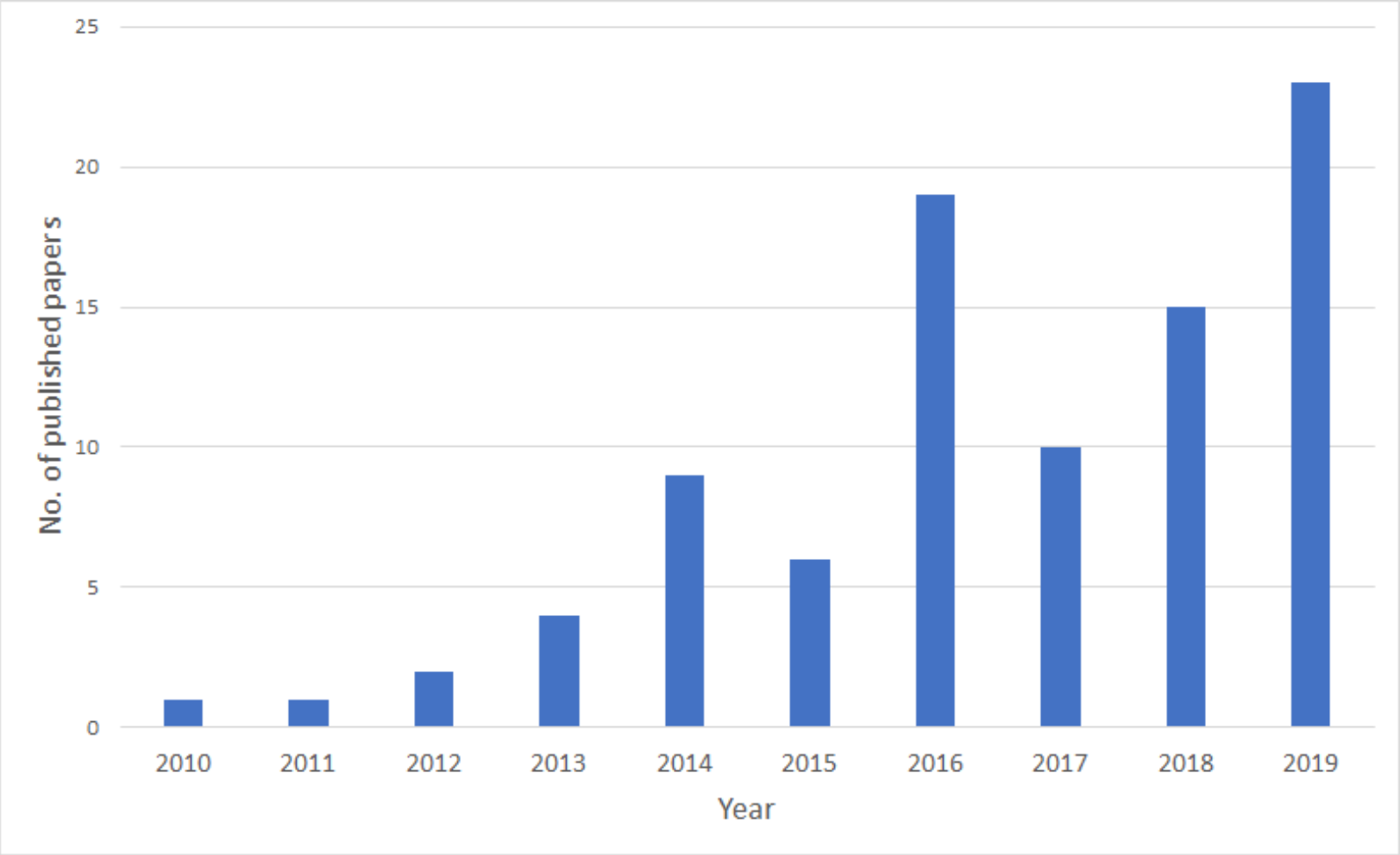

Figure 1: Approximate number of papers published every year on ADI task.

## 2 Arabic Language and Arabic Dialects

'Arabic language' refers to the collection of historically related varieties. These varieties are Modern Standard Arabic (MSA) and informal spoken dialects. MSA is the official language of the Arab World[1] and the primary language of the culture and education system. MSA is mostly written, not spoken. The informal spoken dialects, on the other hand, are the medium of communication in daily life, even on the radio and television shows, from soap operas to music videos. These dialects are primarily spoken, not written, and seen as true native language [16, 2, 4, 6].

The sea of Arabic dialects is vast, with dialects being spoken by more than 300 million native speakers. These dialects can be categorized into various groups based on variety of factors. The most popular factor to categorize the Arabic dialects is based on geographical location. According to the Glottolog [17], a bibliographic database of the world's languages and language families, Arabic dialects are classified into five groups, composing of 38 dialects as shown in Figure 2. These dialectal groups are Arabian Peninsula Arabic, Eastern Arabic, Egyptic Arabic, Levantine Arabic, and North African Arabic. The Glottolog distinguishes the two major dialects of Mesopotamia according to their pronunciation of the letter /q/[2] (ق, qAf) in the word (قلت, qultu,"I said")[3]. The pronunciation of this word in southeastern and central Mesopotamia is /gilit/, while it is pronounced /qeltu/ in northern Mesopotamia.

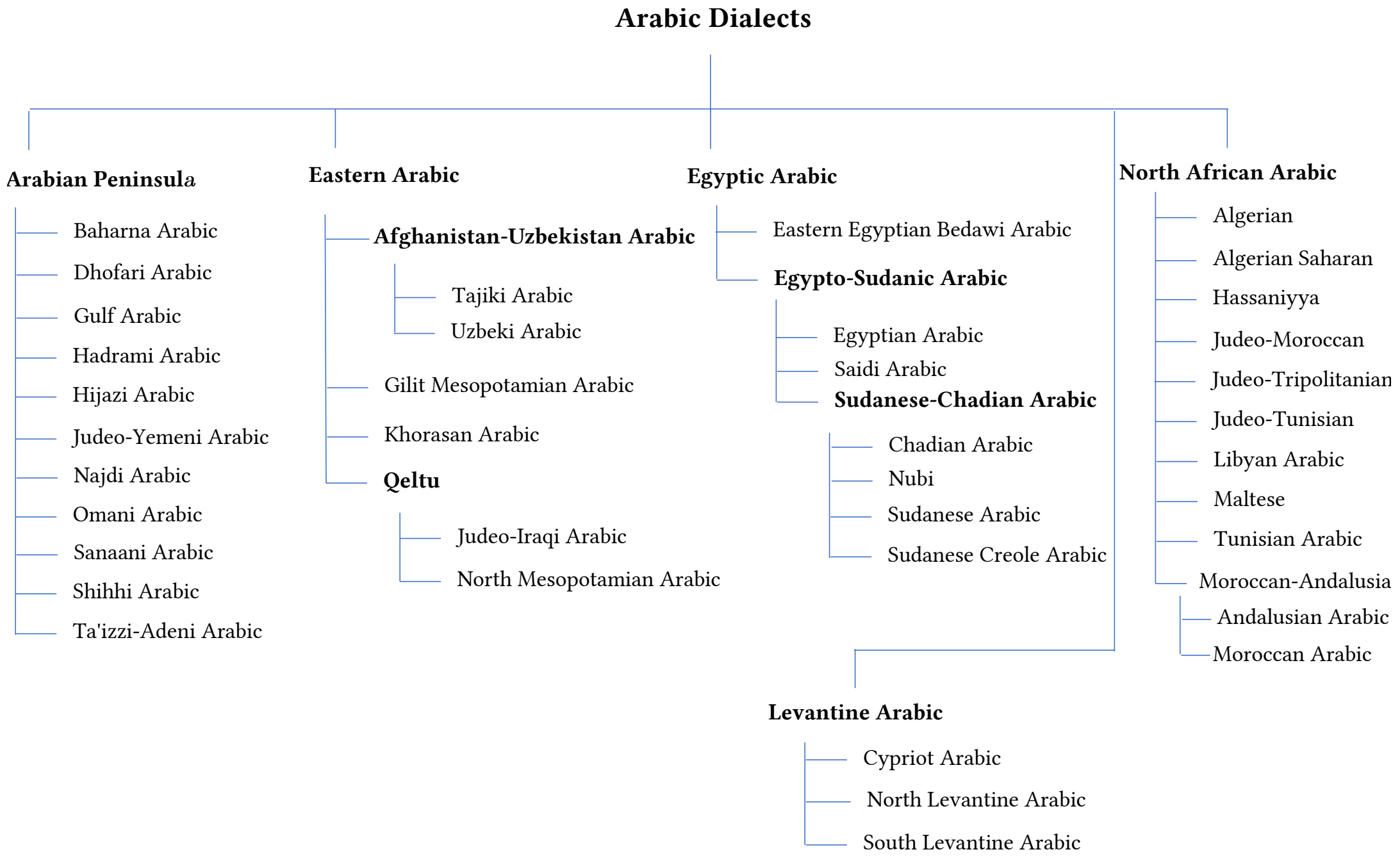

Figure 2: Classification tree of Arabic dialects according to the Glottolog database.

Maltese and Cypriot Arabic are categorized by Borg [18] as isolated "peripheral Arabic" vernaculars due to (a) their geographical and cultural isolation from Arab countries and (b) linguistic acculturation predominantly to one specific foreign language (Cypriot Greek in the case of Cypriot Arabic, Southern Italian in the case of the Maltese). Maltese and Cypriot Arabic also no longer use Arabic script as their writing system [19, 20]. The Nubi, also known as Ki-Nubi, is influenced mainly by the Sudanese dialect and is spoken in Kenya around Kibera, and in Uganda around Bombo. Around 90% of the Nubi lexicon derives from Arabic. The grammar and the pronunciation, however, are simplified versions of Arabic [21, 22, 23, 24]. According to the annual reference on the languages of the world (Ethnologue) in its twenty-third edition [25], there are 36 Arabic dialects as shown in Table 1.

[1]The Arabic speaking countries spread as far west as Morocco and as far east as Oman.

[2]International Phonetic Alphabet (IPA) transcription is used in the paper when discussing the phonology of Arabic words. The transcription is represented in our paper in the form /IPA symbols/.

[3]Throughout the entire paper, Arabic words are represented as follows: (Arabic word, HSB transliteration, "English gloss")

| Region | Dialects | Region | Dialects |
| --- | --- | --- | --- |
| Arabian Peninsula | • Baharna<br>• Dhofari<br>• Gulf<br>• Hadrami<br>• Hijazi<br>• Judeo-Yemeni<br>• Najdi<br>• Omani<br>• Sanaani<br>• Shihhi<br>• Ta'izzi-Adeni | Maghreb | • Algerian<br>• Hassaniyya<br>• Judeo-Moroccan<br>• Judeo-Tripolitanian<br>• Judeo-Tunisian<br>• Libyan<br>• Maltese<br>• Moroccan<br>• Saharan<br>• Tunisian |
| | | Central Asia | • Uzbeki Arabic<br>• Tajiki Arabic |
| Mesopotamia | • Judeo-Iraqi<br>• Mesopotamian<br>• North Mesopotamian | Egypt | • Eastern Egyptian Bedawi<br>• Egyptian<br>• Sa'idi |
| Levant | • Cypriot<br>• North Levantine<br>• South Levantine | Central and Northeast Africa | • Babalia<br>• Chadian<br>• Juba<br>• Sudanese |

Table 1: Arabic dialects in the Ethnologue reference.

Juba Arabic is spoken in southern Sudan and derived from a pidgin[4] based on Sudanese Arabic [26, 27, 28]. The Babalia is a creolized form of Chadian Arabic. The lexicon of Babalia contains 90% Chadian Arabic and 10% Berakuand (the original language of the Babalia people). Babalia Arabic is spoken in southwestern Chad [25].

Until now the primary form of day to day communication in the Arab world has been Arabic dialects, however, The use of dialectal Arabic is evolving due to the increasing prevalence of the Web as a platform for collaborative and community-based sites such as social networking sites, blogs, forums, and reader commentary. This has created a domain where both MSA and Arabic dialects are used relatively equally in written communication [29].

Most Arabic dialects use Arabic script in their writing system. Approximately 22 out of 36 Arabic dialects mentioned in the annual reference “Ethnologue” are currently using Arabic script in the writing process. Arabic script is the second most widely used writing system worldwide after Latin script [30]. There is no distinction between lowercase and uppercase letters in Arabic script. It is written from right to left and consists of 28 letters and 3 different types of diacritical marks: vowel, nunation, and shaddah. Diacritical marks are usually optional. That is, written Arabic text can be diacritized, partially diacritized, or entirely undiacritized [31, 2].

A set consisting of the Latin alphabet, numeric digits, digraph, and symbols like apostrophe is used sometimes informally to write Arabic texts on computers and mobile devices, especially when the keyboard does not support the Arabic script. This is known as Arabish, Araby, Arabizi, and Franco-Arabic [32]. The numeric digits, digraph, and apostrophe are used to represent Arabic phonemes that cannot be represented using Latin script. Arabizi actually gained huge popularity 15 years ago among Arab youth in instant messaging conversations and it is rarely used for lengthy communications [33]. There are many ways to represent one Arabic letter depending on local dialects [32]. Table 2 shows some examples of Arabic letters, which do not have exact equivalent sounds in Latin alphabet, and how they are written in Arabizi system.

### 2.1 Differences between Arabic Varieties

There are a considerable number of differences between MSA and regional dialects. At the same time, regional dialects vary to different degrees in all levels of the language: phonology, morphology, syntax, and lexicon. This section outlines common differences between Arabic dialects and MSA, and characteristics that distinguish certain regional dialects from others.

[4]The term “Pidgin” means a language with a simplified grammars, limited vocabulary, and often drawn from several languages.

| Arabic Letter | IPA | Arabizi Possibilities |
|---|---|---|
| ح | /ħ/ | 7, H, h |
| خ | /x/ | kh, 7', 5 |
| ض | /dˤ/ | 9', d |
| غ | /ɣ/ | 3', gh |

Table 2: Some Arabic Letters with their Arabizi counterparts.

### 2.1.1 Arabic Dialects versus MSA

In the phonological system of the MSA and dialects, there are a considerable number of notable differences [34]:

- The interdental fricatives /θ/ and /ð/ in MSA have been replaced by alveolar stops /t/ and /d/ respectively in some Arabic dialects such as Syrian Arabic. The interdental fricative /θ/ is also sometimes replaced by the alveolar fricative /s/ in Cairene Egyptian Arabic. For example, the Arabic word (كثير, kθyr, "many") is pronounced /kaθiːr/ in MSA, but becomes /ktiːr/ in Syrian dialects. In Egyptian Arabic, the pronunciation of the Arabic word (تمثال, tmθAl, "statue") is /tmsAl/ which corresponds with /tmθAl/ in MSA.
- The glottal stop in MSA words, especially those that appear in the middle of words, is absent and cannot be found in any Arabic dialects. For example, the Arabic word (رأس,rÂs, "head") has the following pronunciation /raʔs/ in MSA. The glottal stop, however, disappears in many Arabic dialects where the pronunciation of the word rÂs becomes /raːs/. Interestingly, the glottal stop is used in some dialects as the reflex of another phoneme. For example, in some Egyptian and Syrian dialects, the /q/ sound becomes /ʔ/.
- There is a tendency in some Arabic dialects to frequently delete short vowels in open syllables [34]. For example, the pronunciation of the word (ثلاثة, θlAθħ, "three") in MSA is /θala:θa/ and becomes /tlaːte/ in Syrian Arabic.
- Many dialects have developed a short central vowel /ə/. This phoneme is a merge of the two MSA vowels /i/ and /u/ in some dialects such as Syrian Arabic. For example, the MSA words (قصة, qSħ, "story") /qisˤæ/ is pronounced in Syrian dialects as /ʔəsˤa/. In other dialects (e.g., Moroccan dialects), the new phoneme /ə/ is a merge of the two vowels /i/ and /a/ [35]. For example, ("he agreed") /wa:faqa/ and ("someone") /waːħid/ in MSA are pronounced as /wafəq/ and /waːħəd/ in Moroccan respectively.

The dialects and MSA also differ morphologically. MSA, in general, has a richer morphology than dialectal Arabic. Contrarily, the cliticization system of the dialects is more complex than that of MSA [6]. Some examples are provided below with explanation:

- Although both MSA and dialectal Arabic have singular and plural forms, dialects mostly lack the dual form that exists in MSA. Specifically, many dialects may preserve the dual form for nouns, but most of them use the plural form instead of the dual form in verb conjugation and pronouns.
- Unlike MSA, many dialects do not have a gender distinction (i.e., masculine and feminine). That is, in MSA, gender is represented morphologically in all word classes, except in first and second persons singular of verbs and pronouns. On the other hand, in many dialects, gender is usually represented in all noun categories in the plural, but not frequently used otherwise as in MSA.
- One of the common features of many dialects is the use of the vowel /i/ and the newly developed short central vowel in some dialects /ə/ instead of /a/ in prefixes of imperfect verbs. For example, /jaktub/ "he writes" correspond to /jiktib/ in Egyptian dialect and /jəktəb/ in Moroccan Arabic.
- The internal passive form (فعل, fçl) /fuʕila/ and (يفعل, yfçl) /yufʕalu/) in many dialects has been replaced by prefixing the verb in the perfective with Ân- or t- as in (انكتب, Ânktb) /enkatəb/ and (تكتب, tktb) /tktəb/ "to be written" in stead of (كُتب>, ktb) /kutiba/.
- Arabic dialects lack case endings, while MSA possesses a complex case system. Case endings are diacritical marks attached to the ends of the word to indicate grammatical function.

- Arabic dialects have a more complex cliticization system than MSA. This complex system allows for circumfix negation, and for attached pronouns to act as indirect objects [6]. Table 3 shows a comparative example of the cliticization system in MSA and the Egyptian dialect.

| | | | | | | | | |
|---|---|---|---|---|---|---|---|---|
| MSA | ولن يكتب له (Three Words) | | | | | | | |
| English Equivalent | And he will not write to him | | | | | | | |
| Egyptian Dialect<br>IPA | وماهيكتبلوش (One Word)<br>/wama:hi:ktblu:ʃ/ | | | | | | | |
| Analysis | Proclitics (3) | | | prefix | verb | Enclitics (3) | | |
| | wa | ma: | h | i: | ktb | l | u: | ʃ |
| | and | not | will | 3person | write | to | him | not |
| Levantine Dialect<br>IPA | وماحيكتبلو (One Word)<br>/wama:ħi:ktblu:/ | | | | | | | |
| Analysis | Proclitics (3) | | | prefix | verb | Enclitics (2) | | |
| | wa | ma: | h | i: | ktb | l | u: | |
| | and | not | will | 3person | write | to | him | |

Table 3: A comparative example of the cliticization system in MSA, Egyptian, and Levantine dialects.

From a syntactic perspective, MSA and Arabic dialects differ in style and sentence composition, as well. For example, MSA and Arabic dialects allow both subject-verb-object (SVO) and verb-subject-object (VSO) word orders, but MSA has a higher rate of VSO sentence occurrence than the dialects [36, 37, 6]. In almost all Arabic dialects, it is very common to have full agreement in number between S and V in all word orders, while MSA has partial agreement in VSO order. Regarding gender agreement, the opposite is true. That is, MSA has full gender agreement for all word orders, while almost all Arabic dialects, especially urban dialects lack gender distinction (e.g., the case of feminine plural subject). Table 4 explains this agreement in number and gender in Arabic varieties by giving examples of sentences written in different orders.

| **Arabic Variants** | **Word Order** | **Agreement** | | **Word Order** | **Agreement** | |
|---|---|---|---|---|---|---|
| | **S-V** | **Number** | **Gender** | **V-S** | **Number** | **Gender** |
| **MSA** | الطلاب كتبوا المقالة<br>The students(M) wrote(they-M) the article. | ✓ | ✓ | كتب الطلاب المقالة<br>The students(M) wrote(he) the article. | ✗ | ✓ |
| | الطالبات كتبن المقالة<br>The students(F) wrote(they-fem.) the article. | ✓ | ✓ | كتبت الطالبات المقالة<br>The students(F) wrote(she) the article. | ✗ | ✓ |
| **Some Gulf Dialects** | الطلاب كتبوا المقالة<br>The students(M) wrote(they-M) the article. | ✓ | ✓ | كتبوا الطلاب المقالة<br>The students(M) wrote(they-M) the article. | ✓ | ✓ |
| | الطالبات كتبوا المقالة<br>The students(F) wrote(they-M) the article. | ✓ | ✗ | كتبوا الطالبات المقالة<br>The students(F) wrote(they-M) the article. | ✓ | ✗ |

Table 4: Examples of sentence word order and agreement in number and gender in Arabic varieties.

In the possessive and genitive constructions, both MSA and dialects have two nouns that can be placed one after the other to indicate possession (e.g., كتاب الولد, "the boy's book"). The term إضافة, ǍDAfħ, "annexation" is used in Arabic grammar to describe the process of placing one noun after another in a genitive construction [38]. Unlike MSA, the dialects have developed an analytic possessive construction, in which a genitive exponent expresses the meaning of possession. Examples of exponents are (تبع tbç, مال mAl, حق Hq) which mean "belonging to" and are used in Levantine and Gulf dialects as in (الكتاب تبع الولد, الكتاب حق الولد "The book belonging to the boy" [34].

Prominent indirect objects, in many Arabic dialects, are cliticized to verbs as in (بدي اكتبلكن رسالة, bidy Aktblkun rsAlh "I want to write you a letter") from Levantine Arabic [39]. Some dialects permit almost any combination of direct and indirect object clitics to be connected with the verb. Therefore, the verbal forms complexity is evidenced by not only the direct and indirect object clitics, but also the negative circumfix (e.g., ما .... ش mA ... š), future particles, and continuous pronominal clitics. Table 3 shows examples of the cliticization system in dialects and how it reduces the number of MSA words in any given phrase and produces one word with many clitics attached to it [34].

Many lexical items found in almost all dialects are actually also found in MSA, but they are used in slightly different ways. That is, these words are used for a specific purpose in MSA, unlike the dialects that ascribe a more general meaning [40, 41]. Table 5 shows some examples of these words. In addition, the newly developed interrogative (question) words in dialects are actually formed from two or more lexical items in MSA. For example, the interrogative words in different Arabic dialects, for (ماذا mAðA ما mA) "What", are ايه Ayh, ايش Ayš, شو šw, شنو šnw. These are varieties of the MSA expression (أي شيء, Ayy šŷ, "which thing").

| **MSA** | **Dialects** |
|---|---|
| جاء بـ *jA' bi* "to come with" | جاب *jAb* "to bring" |
| تشوَّف *tašwwaf* "to observe from above" | شاف *šAf* "to see" |
| سوَّى sawwý "to render equal, arrange" | سوَّ, ساو *sawwa, sAw* "to do, make" |
| راح *rAH* "to go away in the evening" | راح *rAH* "to go away" |

Table 5: Examples of lexical items used in MSA and dialects in slightly different ways.

### 2.1.2 Inter-Dialect Differences

Arabic dialects differ from each other in all language levels. The differences display clearly not only between regional dialects, but also between dialects within the same region and country.

When it comes to phonology, the dialects vary mainly in the pronunciation of the following six consonants: (ج) /ʤ/, (ق) /q/, (ث) /θ/, (ذ) /ð/, (ض) /dˤ/ and (ظ) /ðˤ/. The pronunciation of the MSA phoneme /q/ differs widely from one dialect to another with /g/, /ʔ/, and /q/ being the most common in most forms of the Arabic dialects. Other variations include / ɣ/ in Sudanese and some position of Yemen, /k/ in rural Palestinian, and /ɣ/, or /ʁ/ in some locations in Sudan and Yafi'i dialect in Yemen. The MSA phoneme /ʤ/ is pronounced in different ways such as /ʤ/, /ʒ/, /g/, /j/, or /ɟ/. Regarding the letters (ض) /dˤ/ and (ظ) /ðˤ/, most Arabic dialects do not make a distinction when pronouncing them. For example, they are both pronounced as (ض) /dˤ/ in Coastal Hadhrami and Syrian Arabic dialects, while Gulf pronounces both letters (ظ) /ðˤ/. Table 6 provides examples of some MSA phonemes and their main corresponding pronunciations in Arabic dialects.

Arabic dialects have many differences morphologically. For example, eastern Arabic dialects differ from western Arabic dialects in the first person singular prefix of the imperfect verb (A- in eastern dialects vs. n- in western dialects). For example, the verb (اكتب Aktb, "I write") in most Arabian Peninsula dialects corresponds to (نكتب, nktb)in Moroccan dialects. In addition, the morphological category of mood[5] that exist in MSA for imperfect verbs has disappeared in dialects, which in return have developed a new system of aspectual markers for aspectual distinction. The dialects, however, differ with regard to the verbal form and the scope of the markers. In Egyptian Arabic, for example, the b- marker is used to indicate present habitual and progressive aspect. Table 7 shows the aspectual markers used in different Arabic dialects.

There are some syntactical differences between Arabic dialects. In Cairene Arabic, for example, the question words such as "what", "where", and "when" appear at the end of the question, in contrast to most of other dialects. For instance, the question "What are you doing?" in Cairo will be (بتعمل ايه؟, btçml Ǎyh, "you-do what?") [34]. Also, the question "Who are you?" will be (أنت مين؟, Ǎnta myn?, "you who?"). The variations in word order between dialects also occur in demonstrative articles. That is, the demonstrative articles obligatorily occur in post-nominal position in Egyptian Arabic while other Arabic dialects show both pre- and post-nominal demonstrative constructions. So, "This book" will be (الكتاب ده, AlktAb dh, "book this") in Egyptian dialects. In Levantine dialects, post-nominal and pre-nominal demonstrative articles can occur as follows: (الكتاب هدا, AlktAb hadA, "book this") and (هدا الكتاب, hdA AlktAb, "this book") [16].

Furthermore, vocabulary varies widely between dialects. Not only the same entity can be called different names in different dialects, sometimes the same word can convey a totally different meaning in two different dialects. For example, the word (ماشي, mAŝy) means "OK" in Levantine and Egyptian dialects, but it means "not" in Moroccan dialect. The word (براد, brrAd) means "kettle" in Egyptian dialects, but it means "fridge" in Levantine dialects [16].

[5] Arabic has three types of mood: indicative مرفوع, subjunctive منصوب, and jussive مجزوم

| Letter | MSA (IPA) | Dialectal Main Pronunciations (IPA) | | | |
|---|---|---|---|---|---|
| ق | /q/ | /q/ | /ʔ/ | /g/ | /k/ |
| | | Most of Tunisia, Algeria, and Morocco, Southern and Western Yemen, parts of Oman, Northern Iraq, some parts of the Levant | Most of the Levant, Northern Egypt, some North African towns such as Fez | Most of the Arabian Peninsula, Northern and Eastern Yemen, parts of Oman, Southern Iraq, parts of the Levant, North Africa, Sudan, and Mauritania | Rural Palestinian, some parts of Iraq, and Algeria (Jijel Province) |
| ج | /dʒ/ | /dʒ/ | /ʒ/ | /g/ | /j/ |
| | | Most of the Arabian Peninsula, Algeria, Iraq, Southern Egypt, Sudan, parts of the Levant and Yemen | Most of the Levant and North Africa | Northern Egypt, parts of Oman and Yemen | the Gulf and Southern Iraq and coastal Hadhramaut |
| ث | /θ/ | /θ/ | /t/ | /s/ | |
| | | Most of the Arabian Peninsula, Tunisia, Iraq, parts of Yemen, rural Palestinian, Eastern Libyan, and some rural parts of Algeria | Sudan, most of the Levant, Urban Hejaz, Some words in Egypt, Southern Yemen, Hadhramaut, Morocco, and Algeria | Some words in Egypt, Urban Hejaz | |
| ذ | /ð/ | /ð/ | /d/ | /z/ | |
| | | Most of the Arabian Peninsula, Tunisia, Iraq, parts of Yemen, rural Palestinian, Eastern Libyan, and some rural parts of Algeria | Sudan, most of the Levant, Urban Hejaz, Some words in Egypt, Southern Yemen, Hadhramaut, Morocco, and Algeria | Some words in Egypt, Urban Hejaz | |

Table 6: Examples of the MSA phonemes and their main corresponding pronunciations in Arabic dialects.

| **Dialects** | | **Habitual** | **Progressive/Continuous** | **Future** |
|---|---|---|---|---|
| Syrian | Marker | بـ b- | عم بـ çm b-<br>عم çm | رح *rH*<br>حـ *H-* |
| | Example | بيكتب byktb | عم بيكتب çm byktb<br>عم يكتب çm yktb | رح يكتب *rH yktb*<br>حيكتب *Hyktb* |
| Egyptian | Marker | بـ *b-* | بـ *b-*<br>عمال *çmAl* | حـ *H-*<br>هـ *h-* |
| | Example | بيكتب *byktb* | بيكتب *byktb*<br>عمال يكتب *çmAl yktb* | هيكتب *hyktb*<br>حيكتب *Hyktb* |
| Moroccan | Marker | كـ *k-* | كـ *k-* | غـ *γ-*<br>غادي *γAdy* |
| | Example | كيكتب *kyktb* | كيكتب *kyktb* | غيكتب *γyktb*<br>غادي يكتب *γAdy yktb* |

Table 7: Different Aspectual Markers in Arabic dialects.

# 3 Problem of Arabic Dialect Identification

## 3.1 Definition

Arabic dialect identification in written texts can be defined as the process of building a recognizer that is able to, given an Arabic piece of text T (e.g., sentence, paragraph, document), determine whether or not T was written in dialectal Arabic and in which dialect T was written.

Arabic dialect identification is a most challenging language identification task [6, 42]. That is, language identification can be considered a solved problem, especially when building a system to discriminate between languages that have little features in common. In the case of Arabic dialects, the situation is more complex since the dialects are closely related to each other and share much vocabulary [42]. The following section details the issues and challenges embedded in the task of Arabic dialect identification.

## 3.2 Issues and Challenges

Arabic dialect identification research has yet to overcome numerous challenges associated with discriminating between Arabic dialects; the more new investigations conducted, the more problems will be solved. Some of these difficulties are:

- Arabic speakers mix Arabic varieties in different ways. They may switch between two or more Arabic varieties within the same utterance. This phenomenon is called Linguistic Code Switching (LCS) [43]. The LCS phenomenon is present in the online Arabic content. That is, the writer of online texts sometimes alternates between MSA and at least one Arabic dialect within the same paragraph or even a sentence. It is a very challenging task to identify LCS points whether at a token or sentence level [5, 44].
- The letters of Arabic scripts and the absence of critical marks hide the vocalic and some consonantal differences across dialects. For example, The Arabic letter (ق, qAf) that is pronounced /q/ is pronounced differently across dialects (e.g., /g/,/q/,/ʔ/,/k/,/ʤ/). Relying on the written form of (ق, qAf), we run the risk of mistakenly identifying the writer's dialect, since the written letter in Arabic script does not reveal the writer's pronunciation of it. The same situation is true for the Arabic letter (ظ, Ď) which is pronounced /ðˤ/, but has a variety of pronunciations in different dialects (e.g., /ðˤ/ and /zˤ/). Simply put, the Arabic script may mask the real sounds and pronunciations of letters and words in the written texts of different dialects [45, 46, 39, 47].
- Arabic dialects differ from MSA with regard to the vowel system. The vowel systems also differ substantially from one dialect to another. For example, many short vowels were deleted due to syncope, as in the pronunciation of the MSA word (جبل, jbl, "mountain") /ʤabal/ becomes in Moroccan dialect /ʑbəl/. It is evident that Arabic texts, which are most often written without diacritical marks (i.e., short vowels), do not represent these spoken differences in the vowel system [47, 35]. Moreover, the omission of short vowels results in dialectal words that share the same spelling with MSA words, but mean something entirely different [6]. For example, the Levantine word without dialectical masks (هون, hwn, "here") could be mistaken for a MSA that means (هون, hwn, "make easy").
- There have been many attempts to adapt the Arabic script by adding new letters or dots to the existing letters to account for the new sounds and pronunciations that do not exist in its phonemic inventory [48]. For example, in the written text of the Iraqi dialect, the Persian kAf (گ) has been used to represent the /g/ sound, which is similar to the /g/ in the English word "good". The written Levantine texts usually use the new letter (چ) to represent the pronunciation /g/. In addition, loan words from foreign languages such as English and French have introduced new sounds (e.g., /v/, /p/, /ng/) which led to many attempts to add new letters to the standard Arabic script to represent the precise pronunciation of these letters when they appear in a dialectal Arabic text. This approach of adding new letters for new phonemes to explain the real pronunciations actually creates more orthographic variations between dialects, as shown in Table 8 where the phoneme (e.g., /g/) is transliterated using different letters in different dialects. There are also cases where a letter (e.g., چ) is used to represent different sounds for different dialects [49, 6].
- The majority of the time, Arabic script is used to write dialectal Arabic. Online contents that are generated informally by users, however, are sometimes written in Arabizi, a non-standard romanization consisting of Latin characters, numeric digits, and symbols like the apostrophe. There are various ways in which Arabic letters are transliterated. That is, Arabic letters that do not have similar phonetic approximations in the Latin alphabet are often expressed using numeric digits or a combination of two Latin characters. Arabizi makes Arabic dialect identification based on written texts a challenging task; transliteration in Arabizi follows no standards or rules which creates inconsistency and ambiguity [50, 51].

- A considerable amount of previous research in Arabic NLP has focused on MSA. Therefore, a large number of annotated corpora and freely available resources are built for MSA. Recently, the dialectal Arabic has started to gain the focus of researchers and some resources have been built [6, 52, 1]. The resources available for dialectal Arabic in comparison with MSA, however, let alone resources for other languages such as English, are still severely limited in terms of size and coverage. Therefore, prior research on Arabic NLP that deals with online contents and social media texts have created their own annotated resources to fill in the gap. More efforts are required to built more annotated resources for Arabic dialects in order to develop the computational solutions for dialectal Arabic problems including Arabic dialect identification [6].

| | | Dialects | | |
|---|---|---|---|---|
| | | Iraqi | Levantine | Moroccan |
| phonemes | /g/ | گ | چ | ڭ |
| | /tʃ/ | چ | تش | ڜ |
| | /v/ | ڤ | ڤ | ڥ |
| | /p/ | پ | پ | پ |

Table 8: Attempts to add letters to the Arabic script in dialectal Arabic.

# 4 Arabic Dialect Identification Corpora

ADI research started receiving significant attention approximately a decade ago. The need for annotated corpora for training ADI models and performance evaluation was recognized. Since then, a large number of annotated corpora have been built and made available. A considerable number of the available ADI corpora have been manually annotated in order to increase their reliability. Although the manual approach to corpora annotation results in reliable resources with high quality annotations, this approach requires human effort and a considerable amount of time. The built corpora are usually limited in size, scale, and scope. Therefore, some studies on ADI have applied semi-automatic techniques to annotate the collected dialectal Arabic corpora with less effort.

## 4.1 Manually Annotated ADI Corpora

The following illustrates the ADI corpora with manual annotation. The Arabic Online Commentary (AOC) dataset is one of the earliest ADI corpora publicly available, which designates it a benchmark dataset for later studies. The AOC has been compiled from online commentary by readers of online versions of three Arabic newspapers: AlGhad (Jordanian newspaper), AlRiyadh (Saudi newspaper), and AlYoum AlSabe' (Egyptian newspaper). The AOC dataset contains 108K sentences, of which 63,555 sentences are MSA, and the remaining represent three dialects: Egyptian, Gulf, and Levantine. The annotation process has been conducted by Amazon's Mechanical Turk (MTurk) workers. For each sentence, they indicate the level of dialectal Arabic, and which dialect it is [11]. The AOC dataset has been widely used as a benchmark dataset to evaluate various ADI techniques in the literature and to compare their performances [5, 44, 53, 54, 55, 56, 57, 58, 59]. Following the work of Zaidan and Callison-Burch [11], Cotterell and Callison-Burch [53] built Extended AOC dataset consisting of 27,239 user comments from online newspapers. However, the corpus covered two more dialects (Maghrebi, and Iraqi dialects) in addition to the three covered by Zaidan and Callison-Burch [11] (Levantine, Gulf, and Egyptian). They also created Twitter corpus consisting of 40,229 tweets from the five aforementioned dialects. The data of the corpora was manually collected and annotated by workers from MTurk. Additionally, McNeil and Faiza [60] created the Tunisian Arabic Corpus (TAC), consisting of 895,000 words collected from three sources: (a) traditional written sources such as song lyrics, collections of proverbs, folklore, folk poetry, television shows, plays, and movie screenplays, (b) new written sources such as blogs and discussion forums, (c) transcriptions of audio sources such as radio broadcasts and podcasts. The TAC data was collected, identified, checked, and transcribed when required by Tunisian college-level students and workers from MTurk. The corpus is only publicly available through a web-based interface in which a user can search for a word. It is not publicly available for downloading.

Parallel corpora for dialectal Arabic have also been the focus of many studies. Zbib [10] created Levantine-English and Egyptian-English parallel corpora, consisting of 1.1M words and 380K words, respectively. The dialectal Arabic sentences were collected originally from the Web (largely from online user groups and weblog) and then filtered using a set of dialectal words to remove MSA and non-dialectal sentences. MTurk workers were responsible for the classification of each sentence into Levantine or Egyptian dialect and the translation of these sentences into English. The Multidialectal Parallel Corpus of Arabic (MPCA) contains 2,000 parallel sentences, covering five Arabic dialects: Egyptian, Tunisian, Jordanian, Palestinian, and Syrian Arabic. The Egyptian sentences were selected from the Egyptian portion of the Egyptian-English corpus built by Zbib et al. [10]. Then, non-professional translators hired on MTurk were asked to translate the Egyptian sentences into their native dialect [12]. The MPCA corpus is publicly available upon request. Sadat et al., [61] built a Social Media corpus collected from blogs and forums. The corpus, manually annotated, contains 61,859 sentences and is publicly available. It covers 18 country-level dialects spoken in Algeria, Bahrain, Egypt, Emirates, Iraq, Jordon, Kuwait, Lebanon, Libya, Mauritania, Morocco, Oman, Palestine, Qatar, Saudi Arabia, Sudan, Syria, and Tunisia.

The Dial2MSA parallel corpus created by Mubarak [62] contains dialectal Arabic tweets in four main Arabic dialects (Egyptian, Maghrebi, Levantine, and Gulf) and their corresponding MSA translations. The tweets were first collected using Twitter API, and then filtered using a set of distinctive words for each dialect. The crowdsourcing platform (CrowdFlower) was then used to hire native speakers of each dialect in order to translate each tweet into its corresponding MSA. The final corpus contains 16,000 pairs for Egyptian-MSA, 8,000 pairs for Maghrebi-MSA, and 18,000 pairs for each Gulf-MSA and Levantine-MSA. Furthermore, the PADIC, a Parallel Arabic DIalect Corpus, was created from recorded conversations from everyday life, movies, and TV shows of Annaba's dialect, spoken in the east of Algeria, and Algiers's dialect. The PADIC corpus is publicly available. The sentences were manually transcribed and translated by native speakers into MSA as well as three more Arabic dialects: Sfax's dialect spoken in the south of Tunisia, Syrian and Palestinian dialects. The total number of parallel sentences in the corpus is 6,400. The MADAR travel domain corpus presented by Bouamor et al. [63] was used in the MADAR shared task on Arabic fine-grained dialect identification. More specifically, it is used in the MADAR travel domain Arabic dialect identification subtask. The MADAR travel domain corpus is a large-scale collection of parallel sentences covering the dialects of 25 Arab cities, in addition to English, French and MSA. The MADAR corpus is a commissioned translation of selected sentences from the Basic Traveling Expression Corpus (BTEC) [64] in English and French to the dialects of 25 Arab cities. The BTEC is a multilingual spoken language corpus containing tourism-related sentences. The MADAR corpus consists of two parts: the first called "Corpus-26" has 2,000 sentences translated into 25 Arab city dialects in parallel as well as MSA , and the second one called "Corpus-6" consists of 10,000 additional new sentences from the BTEC corpus translated into only five selected Arab city dialects, plus MSA.

The MADAR Twitter corpus was used in the MADAR Twitter user dialect identification subtask that was organized as part of the MADAR shared task on Arabic fine-grained dialect identification [15]. This corpus contains 2,980 Twitter user profiles from 21 different countries. The Twitter corpus was used to evaluate the participating systems in the MADAR Twitter user dialect identification subtask. The Gumar corpus is a large-scale corpus of Gulf Arabic consisting of 1,236 forum novels with about 112 million words. The corpus was manually annotated at the document level for sub-dialect information (i.e., dialects spoken in the six countries of the Gulf Cooperation Council: Bahrain, United Arab Emirates, Kuwait, Saudi Arabia, Oman, or Qatar).

Elfardy and Diab [43] built the Linguistic Code Switching (LCS) corpus collected from forum posts and manually annotated at token-level for linguistic code switching between MSA and three Arabic dialects (Egyptian, Gulf, and Levantine). The corpus contains 1,170 forum posts corresponding to a total of 27,173 tokens. This corpus is the first of its kind created for dialectal Arabic identification in code-switched data. That is, native speakers of Arabic usually mix dialectal Arabic and Modern Standard Arabic in the same utterance. The dialectal Arabic identification in code-switched data aims to identify the points at which the language switches from MSA to a dialectal Arabic and vice versa. The LICSD'2014 and LICSD'2016 are two datasets manually annotated at token-level and provided by the shared tasks for Language Identification in Code Switched Data (LICSD) in 2014 and 2016 [65, 66]. The LICSD-2014 includes code-switched data from four language pairs: Modern Standard Arabic-Egyptian dialect (MSA-EGY), Nepali-English (NEPEN), Mandarin-English (MAN-EN), and Spanish-English (SPA-EN). The data for the MSA-EGY variety pair was compiled from Twitter and online reader commentaries. They harvested 9,947 tweets and 6,723 commentaries (half MSA and half Egyptian) from an Egyptian newspaper provided by the Arabic Online Commentary (AOC) Dataset. The LICSD-16 dataset created by Molina et al. [66] includes code-switched data from two language pairs: English-Spanish, and MSA-EGY. For MSA-EGY, they published 11,241 tweets (8,862 tweets for the training set, and 1,117 tweets for the development set, 1,262 tweets for the test set). Many studies in ADI literature have used The LICSD-2014 and LICSD-2016 corpora to evaluate their token-level ADI systems [67, 55, 68].

The VarDial'2016 ADI corpus was released by the organizers of the Discriminating between Similar Languages (DSL) shared task in the VarDial'2016 workshop [69]. The shared task offered two subtasks. The subtask 2 managed the Arabic dialect identification in speech transcripts. The utilized corpus is based on Arabic transcribed speeches presented by Ali et al. [70], covering four Arabic dialects: Egyptian, Gulf, Levantine, and North African, as well as MSA. Malmasi et al. [69] released 7,619 sentences for training and development and 1,540 sentences for evaluation. The corpus has been used for evaluation by all the systems submitted to the ADI task [69]. The VarDial'2017 ADI corpus was provided for the second edition of the ADI shared task of the 2017 VarDial Evaluation Campaign on Natural Language Processing (NLP) for Similar Languages, Varieties and Dialects [71]. It is similar to the VarDial'2016 corpus as they are both based on speech transcripts and cover the same Arabic dialects. However, the VarDial'2017 dataset provides acoustic features to the task participants [71]. The VarDial'2017 corpus contains 13,825 samples, the development set contains 1,524 samples, and the test contains another 1,492 samples. It was used to evaluate the ADI systems that participated in the campaign. Then, organizers of the VarDial Evaluation Campaign 2018 released the third edition of the ADI task VarDial'2018 ADI corpus [72]. Like the previous two editions of the ADI task, the VarDial'2018 included four Arabic dialects: Egyptian, Levantine, Gulf, North African, as well as MSA. In addition, the data released for training and development were the same data released in the 2017 edition of the ADI task. Regarding testing, two datasets were prepared: an in-domain test set from broadcast news and an out-of-domain dataset from YouTube, containing 5,345 samples. The two test sets were merged. Moreover, the participants were provided phonetic, acoustic, and lexical features. The corpus was used to evaluate the systems submitted to the 2018 ADI shared task. It was also used by the study of Dinu et al. [73] to evaluate their technique for ADI.

### 4.2 Semi and Fully Automatically Annotated ADI Corpora

Huang utilized semi-supervised learning for automatic ADI annotations [56]. The study examined two semi-supervised methods: self-training and co-training. In self-training, a strongly supervised classifier trained on a small amount of gold labeled data (AOC corpus) was used to label a large amount of data (646M words) extracted from Facebook posts. The co-training method employed two classifiers to annotate the same unlabeled data. Then, only the sentences on which the two classifiers agree (they have the same predictions) were added to the final corpus (476M words).

Some studies classify the collected data into a set of dialects using a list of distinctive regional words and phrases [74, 75]. These words are considered “seed words” in the process of automatically collecting and annotating the data for each dialect. Almeman and Lee [74] used 1,043 dialectal distinctive words for five Arabic dialects to automatically collect and annotate data for each dialect from the Web using Bing API. In addition to the use of distinctive regional words, the geographical locations provided in users' profiles on many social media sites such as Facebook, Twitter, and YouTube were also used to automatically classify the collected data from these social networks [76, 77, 78, 79, 80, 81]. Table 9 lists the existing manually, semi- and fully automatically annotated ADI corpora.

## 5 Evaluation Metrics

Clearly, the Arabic dialect identification task presents a challenge to categorize items (e.g., tokens, sentences, paragraphs, documents) into a set of Arabic dialects. The commonly used evaluation metrics for text classification systems are precision, recall, and F1-score [87]. In the field of text classification, precision and recall are defined in terms of correctly classified positive instances. Thus, precision is the percentage of correctly classified positive instances among the total classified positive instances, and recall is the percentage of correctly classified positive instances among all positive instances examined [87, 88]. Moreover, F1-score is the weighted harmonic mean of precision and recall:

$$F_1 = 2.\frac{Precision.Recall}{Precision + Recall}$$

Table 9: Summary of Arabic Dialect Identification (ADI) Corpora

| Corpus | Labels | Size | Annotation Level | Annotation Method | Source |
|---|---|---|---|---|---|
| Arabic Online Commentary (AOC) dataset [11] | Levantine, Gulf, Egyptian, MSA | 108K sentences | Sentence-level annotated corpus | Annotation was performed manually via crowdsourcing | Reader commentary from three online Arabic newspapers: Al-Ghad, Al-Riyadh, Al-Youm Al-Sabe’ |
| Extended AOC dataset [53] | Egyptian, Gulf, Levantine, Maghrebi, Iraqi | 27,239 newspapers’ reader comments | Comment-level annotated corpus | Annotation was performed manually via crowdsourcing | Reader commentary from online five Arabic newspapers: Al-Youm Al-Sabe’, Al-Riyadh, Al-Ghad, Ech Chorouk El Youmi, Al-Wefaq. |
| Twitter corpus [53] | Egyptian, Gulf, Levantine, Maghrebi, Iraqi | 40,229 tweets | Tweet-level annotated corpus | Annotation was performed manually via crowdsourcing | Twitter |
| Tunisian Arabic Corpus (TAC) [60] | Tunisian dialect | 895,000 words | One corpus of Tunisian dialect | TAC was manually identified, checked, and transcribed by Tunisian students and via crowdsourcing | 1. Traditional written sources (e.g., folklore) 2. New written sources (e.g., online discussion forums) 3. Transcription of audio sources (e.g., podcasts) |
| Levantine/English, Egyptian/English parallel corpora [10] | Levantine, Egyptian, English | 1.1M words for Levantine/English, 380K words for Egyptian/English | Sentence-level parallel corpus | Three steps (dialect classification, sentence segmentation, and translation from dialectal Arabic into English) were performed manually via crowdsourcing | Large data of monolingual Arabic text harvested from the Web (mainly from weblog and online user groups) by the LDC [82] |
| Multidialectal Parallel Corpus of Arabic (MPCA) [12] | Egyptian, Tunisian, Jordanian, Palestinian, Syrian, MSA, English | 2,000 parallel sentences | Sentence-level parallel corpus | Four native speakers were asked to manually translate 2,000 sentences written in Egyptian into their dialects. A fifth translator was asked to translate the same sentences into MSA. The sentences were translated into English via crowdsourcing | Egyptian portion of the Egyptian/English parallel corpus built by Zbib et al. [10] |
| Social Media dataset [61] | Algeria, Bahrain, Egypt, Emirates, Iraq, Jordan, Kuwait, Lebanon, Libya, Mauritania, Morocco, Oman, Palestine, Qatar, Saudi Arabia, Sudan, Syria, Tunisia | 61,859 sentences | Sentence-level annotated corpus | Texts were manually collected, segmented into coherent sentences and then classified into 18 dialects | Blogs and forums of different Arabic-speaking countries |

Table 9: Continued

| Corpus | Labels | Size | Annotation Level | Annotation Method | Source |
|---|---|---|---|---|---|
| Dial2MSA parallel corpus [62] | Egyptian, Maghrebi, Levantine, Gulf, MSA | 16,000 pairs for Egyptian/MSA, 8,000 pairs for Maghrebi/MSA, 18,000 pairs for Levantine/MSA, 18,000 pairs for Gulf/MSA | Tweet-level parallel corpus | Tweets were manually translated into MSA via crowdsourcing. The translations obtained for each tweet were then verified manually via crowdsourcing to verify whether each pair (i.e., Dialect/MSA) share the same meaning | Twitter |
| PADIC [83] | Annaba's dialect, Algiers's dialect, Sfax's dialect, Syrian, Palestinian, MSA | 6,400 parallel sentences | Sentence-level parallel corpus | Recordings for Annaba's dialect and Algiers's dialect were manually transcribed. Then, each of the two corpora were manually translated into the other. Finally, the text results were translated (also by native speakers) into MSA, Syrian, Palestinian, and Sfax's dialects | Recorded Conversations from everyday life, movies and TV shows |
| MADAR travel domain corpus, Corpus-26 & Corpus-6 [63] | MSA and 25 Arabic city dialects (Rabat, Fes, Algiers, Tunis, Sfax, Benghazi, Tripoli, Alexandria, Cairo, Aswan, Khartoum, Beirut, Damascus, Aleppo, Jerusalem, Amman, Salt, Baghdad, Mosul, Basra, Doha, Muscat, Riyadh, Jeddah, Sana'a) | 2,000 parallel sentences for MSA and the 25 city dialects, 12,000 parallel sentences for MSA and the five selected cities (Doha, Beirut, Cairo, Tunis, and Rabat) | Sentence-level annotated corpus | Manual translation from English into MSA and 25 Arabic city dialects | Selected sentences from the Basic Traveling Expression Corpus (BTEC) created by [64] and written in English and French |
| MADAR Twitter Corpus [15] | Algeria, Bahrain, Djibouti, Egypt, Emirates, Iraq, Jordan, Kuwait, Lebanon, Libya, Mauritania, Morocco, Oman, Palestine, Qatar, Saudi Arabia, Somalia, Sudan, Syria, Tunisia, Yemen | 2,980 Twitter user profiles | Document-level annotated corpus | Manual annotation was performed by native speakers | Twitter |

Table 9: Continued

| Corpus | Labels | Size | Annotation Level | Annotation Method | Source |
|---|---|---|---|---|---|
| Gumar corpus [84] | Saudi Arabia, United Arab Emirate, Kuwait, Oman, Qatar, Bahrain, Gulf (other), Arabic (other) | 1,236 documents | Document-level annotated corpus | Manual annotation by native speakers | MS Word documents of novels in an online forum |
| MSA/DA Linguistic Code Switching corpus [43] | Egyptian, Gulf, Levantine, MSA | 1,170 forum posts (27,173 tokens) | Token-level annotated corpus (context sensitive and context insensitive annotation) | Annotation was performed manually by native speakers | Egyptian and Levantine forums |
| Language Identification in Code Switching Data (LICSD'2014) [65] | MSA/Egyptian corpus | 9,947 tweets and 6,723 commentaries | Token-level annotated corpus | Annotation was performed manually by native speakers | Twitter and online reader commentaries from an Egyptian newspaper provided by the Arabic Online Commentary (AOC) Dataset (Zaidan and Callison-Burch [11]) |
| Language Identification in Code Switching Data (LICSD'2016) [66] | MSA/Egyptian corpus | 11,241 tweets | Token-level annotated corpus | Annotation was performed manually by native speakers | Twitter |
| VarDial'2016 ADI corpus [69] | Egyptian, Gulf, Levantine, North African, MSA | 9,159 sentences | Sentence-level annotated corpus | Dataset was originally compiled by Ali et al. [70], containing transcribed speech in MSA and in four dialects. Malmasi et al. [69] extracted text from twelve hours of speech per dialect for training and testing | Multi-dialectal speech corpus created from high-quality broadcast, discussion programs, and debate |
| VarDial'2017 ADI corpus [71] | Egyptian, Gulf, Levantine, North African, MSA | 16,841 sentences | Sentence-level annotated corpus with lexical and acoustic features | Dataset was originally compiled by Ali et al. [70] where the text transcriptions were automatically generated using large-vocabulary speech recognition (LVCSR). The dataset contains transcribed speech in MSA and in four dialects | Multi-dialectal speech corpus created from high-quality broadcast, discussion programs, and debate |

Table 9: Continued

| Corpus | Labels | Size | Annotation Level | Annotation Method | Source |
|---|---|---|---|---|---|
| VarDial'2018 ADI corpus | Egyptian, Gulf, Levantine, North African, MSA | 22,186 sentences | Sentence-level annotated corpus with acoustic features and phonetic inputs | Dataset was originally compiled by Ali et al. [70] where the text transcriptions were automatically generated using large-vocabulary speech recognition (LVCSR). The dataset contains transcribed speech in MSA and in four dialects | Multi-dialectal speech corpus created from high-quality broadcast, discussion programs, and debate |
| DART [85] | Egyptian, Maghrebi, Levantine, Gulf, Iraqi | 24,280 tweets | Tweet-level annotated corpus | Annotation was performed manually via crowdsourcing | Twitter |
| Arap-Tweet corpus [86] | Morocco, Algeria, Tunisia, Libya, Egypt, Sudan, North Levant, South Levant, Iraq, Gulf, Yemen | 1,100 user profiles with their 2.4M Tweets corpus | Tweet-level annotated corpus | Annotation was performed manually by experienced annotators | Twitter |
| Arabic Multi-dialect text corpora [74] | Gulf, Levantine, Egyptian, North African | 48M words in total for the four corpora | One corpus for each dialect | Bootstrapping the Web using Bing API and 1,043 distinctive words and phrases for the four dialects | Web texts |
| YouTube Dialectal Arabic Commentary Corpus (YouDACC) [76] | Egyptian, Gulf, Iraqi, Maghrebi, Levantine | 630,817 sentences | Sentence-level annotated | Annotation was performed semi-automatically based on a list of keywords for each dialect, Youtube API, and geographical locations of the authors of the retrieved comments | User comments on YouTube videos |
| Shami dialects corpus (SDC) [80] | Syrian, Lebanese, Jordanian, Palestinian | 117,805 sentences | Sentence-level annotated corpus | Semi-automatic annotation for collected tweets using only Twitter API and geographical location, manual annotation for collected contents from the Web | Twitter, discussion forums, and online blogs for public Levantine figures |
| Twitter multi-dialectal corpus of Arabic [78] | Saudi Arabia, Egypt, Kuwait, United Arab Emirate, Qatar, Other (such as Morocco, Syria, Lebanon, Algeria and Sudan) | 6.5M tweets | Tweet-level annotated corpus | Annotation was performed semi-automatically based on distinctive keywords of dialects and the location information of the tweets' authors from their profiles | Twitter |

Table 9: Continued

| Corpus | Labels | Size | Annotation Level | Annotation Method | Source |
|---|---|---|---|---|---|
| Tweets corpus [81] | 29 City-level dialects: Alexandria, Cairo, Giza, Baghdad, Karbala, Zubair, Amman, Aqaba, Irbid, Ahmadi, Hawally, Kuwait City, Muscat, Salalah, Sohar, Gaza, Nablus, Ramallah, Al-Rayyan, Doha, Dammam, Jeddah, Riyadh, Abu Dhabi, Al Ain, Dubai, Aden, Sana, Taiz | 1/4 billion tweets | Tweet-level annotated corpus | Annotation was performed automatically using python geocoding libaray as well as a third-party geocoder to acquire location labels on the data | Twitter |
| Semi-supervised labeled corpus [56] | Egyptian, Gulf, Levantine, MSA | 476M words for co-training and 646M for self-training | Sentence-level annotated corpus | Semi-supervised learning (co-training method) | Facebook |
| Weakly labeled data [56] | Egyptian, Gulf, Levantine, MSA | 66M words | Sentence-level annotated corpus | Classification of collected sentences from Facebook was performed automatically based on the authors' country information from their profiles | Facebook |
| Tunisian dialect electronic writing [75] | Tunisian dialect (Latin transcription) | 43,222 messages | Message-level annotated corpus | Lexicon-based classification method, six lexicons: TLD (Tunisian words written in Latin characters), English, English-SMS, French, French-SMS, Emoticons | SMS messages, Tunisian forums, sites, and Facebook |
| Tunisian dialect corpus written in Latin alphabet (TLD), Tunisian dialect corpus written in Arabic alphabet (TAD) [77] | Tunisian dialect (Latin and Arabic transcriptions) | 31,158 messages for TLD and 7,145 messages for TAD | Message-level annotated corpus | Lexicon-based classification method, five lexicons: TAD (Tunisian words written in Arabic alphabet), TLD (Tunisian words written in Latin alphabet), MSA, French, English | Facebook |
| Twitter Arabic Dialect Corpus [79] | Gulf, Iraqi, Egyptian, Levantine, North African | 210,915 tweets | Tweet-level annotated corpus | Classification of the collected tweets was performed semi-automatically based on the use of distinguished words as well as the geographical locations of the tweets' authors | Twitter |

Many studies carried out for the ADI task have assessed the performance of their implemented models using theses three common evaluation metrics: precision, recall, and F1-score [11, 5, 89, 53, 54, 90].

The use of precision, recall and F1-score are arguably flawed, because they neglect the true negative (i.e. correctly predicting that an instance is not among the classified positive instances). Therefore, accuracy is a recommended evaluation metric for classification problems in addition to precision, recall, and F1-score [88, 91]. Accuracy is defined as the percentage of correctly classified instances (both true positives and true negatives) among the total number of instances examined. The accuracy score is used as a performance measure for the ADI task in some studies [11, 5, 92, 54, 55, 56]. Although accuracy is considered a sufficient measure for evaluating performance in classification problems, it is only effective when there is a symmetric dataset in which false positives (i.e. falsely predicting that an instance is among the classified positive instances) and false negatives (i.e. falsely predicting that an instance is not among the classified positive instances) are almost the same. Therefore, most studies do not only use accuracy as a measure but also precision, recall, and F1-score.

In the first edition of the ADI subtask, which was part of the Discriminating between Similar Languages (DSL) shared task, the organizers used a macro-averaged F1-score as an official score to evaluate participating systems [69]. The macro-averaged metric calculates the metric for each Arabic dialect independently and then takes the average. The second edition of the ADI shared task within the VarDial Evaluation Campaign 2017 used the weighted F1-score as a main evaluation metric. In other words, the participating systems were ranked based on their weighted F1-score (weighted by the number of examples in each class) [71]. The organizers of the MADAR shared task on Arabic dialect identification in 2019 ranked the participating systems based on macro-averaged F1 scores and also reported the performance in terms of macro-averaged precision, macro-averaged recall and accuracy [15].

## 6 Arabic Dialect Identification Approaches

The field of Arabic dialect identification is relatively new, having caught the attention of researchers starting in 2011. At that time, the most serious obstacles were the lack of NLP tools and annotated resources for dialectal Arabic in general, and even more so for the ADI task. Therefore, the earliest studies, employed dictionaries and rules to distinguish Arabic varieties. The language modeling approach was also extensively examined for ADI. This approach involves assigning probabilities to sentences in a language as well as assigning a probability to each sequence of words. It also assigns a probability for the likelihood of a given word to follow a sequence of words [93] where word n-grams indicates the sequences of n consecutive words and character n-grams indicates the sequences of n consecutive characters. Byway of this approach, various orders of n-gram language models have been scrutinized to identify Arabic dialects. Since then, many studies have developed their own manually annotated ADI corpora, allowing for “supervised” Machine Learning (ML) methods and feature engineering to be employed to identify dialectal Arabic. Currently, deep learning is considered state-of-the-art machine learning, and research is being conducted on the use of deep learning for ADI. These experiments build several neural network architectures with embedded representations of text units used as input features. Multiple approaches are now being combined to create high performance Arabic dialect identification systems.

Arabic dialect identification approaches are categorized as follows: (a) unsupervised approaches, (b) semi-supervised approaches, (c) feature engineering supervised approaches, (d) deep supervised approaches. Presented below are detailed descriptions of each category’s main techniques specifically designed for Arabic dialect identification.

### 6.1 Language Modeling and Minimally Supervised Approaches

A decade ago when the ADI task started to emerge, the annotated resources in terms of size and coverage of Arabic dialects were sparse. Therefore, early works on ADI relied on using dictionaries, rules, and language modeling. They also depended on the few morphological analyzers to handle Out Of Vocabulary (OOV) words. The majority of early works mainly utilized a combination of methods in which each method should assign a score to each word in the test texts and the final decision rely on the aggregate score.

One of the early studies carried out by Elfardy and Diab [43] addressed automatic identification of token-level dialectal words (LCS points) in Arabic social media texts in which native Arabic speakers frequently mix dialectal Arabic and MSA. Elfardy and Diab considered two settings when performing the experiments on token-level ADI: (a) context sensitive, context is accounted for when considering the class label of each word in the text, and (b) context insensitive. Their study utilized a set of rules, dictionaries, language models, and a MSA morphological analyzer to tackle the problem. They used Tharwa, a three way DA-MSA-English machine readable dictionary collected from paper dictionaries and other resources obtained from the Linguistic Data Consortium (LDC) and created by Diab et al., [52]. A set of rules was used to model the possible phonological varieties of each word in various dialects. The MSA morphological analyzer, ALMOR [31], was also employed by Elfardy and Diab [43] to distinguish between MSA and dialectal Arabic words. To explain, every word achieves a score based on ALMOR's ability to analyze the word, and the word's "existence" in the dictionary. Regarding the Language Model (LM), they built three 3-gram LMs using the SRILM toolkit [94]. These LMs include (a) MSA-LM for MSA and (b) DA-LM for dialectal Arabic and (c) MSA-DA LM that incorporates the entries in both MSA-LM and DA-LM. Next, the LMs and their probabilities for n-garms were used to calculate the DA and MSA scores for each n-gram. Finally, they combined all the scores from all the aforementioned resources and determined each word's dialect based on the highest aggregate score. The evaluation corpus of 1,170 forum posts was manually collected and annotated. Using only the Language Modeling approach, their ADI system achieved best performance of F1-score=84.9% with context-insensitive setting. On the other hand, their system that relied on the aggregate score produced by combining methods (dictionaries, rules, MSA morphological analysis, and language modeling) achieved the best result of F1-score=72.4% in the experimental setting where context was taken into account. Additionally, Elfardy et al. [5] proposed a system to perform automatic identification of linguistic code switches in Arabic. The system relied on the use of language models that passed Out of Vocabulary (OOV) words to morphological analyzer CALIMA designed for Egyptian and MSA Arabic [95] in which OOV words were tagged "MSA," "Egyptian," or "both," depending on their analyses. The system achieved an F1-Score of 76.5% when taking context into account.

Moreover, other early works on ADI examined language modeling methods by investigating the use of various n-gram based features at both character-level and word-level to identify dialectal Arabic [11, 6]. The studies that used language modeling for ADI confirmed the simplicity and efficiency of this approach. Furthermore, Zaidan and Callison-Burch [11] built word trigram models for Levantine, Gulf, Egyptian, and MSA. Their method scores test sentences with all models, and the Arabic variety of the model assigning the highest score (i.e. lowest perplexity) is chosen. The 2-way classification scenario of their method (MSA vs. dialects) achieved 77.8% accuracy on the AOC dataset. Their method excelled at distinguishing the dialects from each other (Levantine vs. Egyptian vs. Gulf), reaching 83.5% accuracy. In another study, Zaidan and Callison-Burch [6] explored higher-order character language models as well as word models. They tried character unigram, trigram & 5-gram LMs as well as word unigram, bigram & trigram LMs. They concluded that a unigram word LM performs best on the AOC dataset with an accuracy of 85.7% while the character 5-gram LM fell slightly behind with an accuracy of 85.0%. The Prediction by partial matching (PPM) technique has been utilized to identify Arabic dialects in a number of studies [89, 96, 97]. Lippincott et al. [97] participated in the MADAR shared task on fine-grained Arabic dialect identification [15], task 2 addressing Twitter user dialect identification. They They created a character-based model using a PPM language modeling technique. They experimented with various values of maximal order (N) to determine the likelihood of observing a symbol following a given context of up to N characters. Their experiments indicated that $N = 3$ was the best value for Arabic dialect identification. They achieved an F1-score of 50.43% and was ranked 6th out of the 9 participating systems.

Huang [56] tested the eligibility of semi-supervised learning techniques, self-training and co-training, for the ADI task of obtaining more annotated data in the training phase. The self-training technique applies a strongly supervised classifier, which was trained on gold standard data, on the unlabeled data to obtain more annotated data. Huang [56] employed a strongly supervised classifier trained on the AOC training dataset to automatically annotate a large unlabeled dataset collected from social media (646M words). The additional annotated data was used to train a new classifier, achieving 84.4% accuracy on the AOC test dataset and 65.5% on Facebook data. Surface features (i.e., word unigram information) were used to train the classifier. The study focused on simple features, since it intended to examine the efficiency of the semi-supervised techniques rather than examining the features themselves. In contrast, the co-training applies two classifiers to automatically annotate unlabeled data. Then, only sentences on which the two classifiers agree are used to train a new classifier. Huang [56] implemented two classifiers: (a) a strongly supervised classifier trained on the AOC training dataset, and (b) a weakly supervised classifier trained on automatically annotated Facebook data based on the country indicated on the author's Facebook profile. The co-training classifier achieved 86.2% accuracy on the AOC test dataset and 67.7% on Facebook data. Table 10 presents a summary of the Arabic Dialect Identification (ADI) models developed based on unsupervised and semi-supervised approaches.

Table 10: Summary of Arabic Dialect Identification (ADI) Models based on Language Modeling and Minimally Supervised Methods

| Reference | Learning Method/Model | Features Used | Processing Level | Class Labels | Evaluation Corpus | Evaluation Results |
|---|---|---|---|---|---|---|
| Elfardy and Diab [43] | Language modeling | Word 3-gram LMs | Token level (LCS points) | Egyptian, Gulf, Levantine, MSA | Their own manually annotated dialectal Arabic forum posts (context insensitive annotation) | F1-score=84.9% |
| Elfardy and Diab [43] | Dictionaries, rules, MSA morphological analysis, and language modeling | Word 3-gram LMs, MSA morphological analyzer (ALMOR) scores, rules based on phonological variations between dialects | Token level (LCS points) | Egyptian, Gulf, Levantine, MSA | Their own manually annotated dialectal Arabic forum posts (context sensitive annotation) | F1-score=72.4% |
| Elfardy et al. [5] | Language modeling with a back off to a MSA and Egyptian morphological analyzer | Word 5-gram LM, MSA and Egyptian morphological analyzer (CALIMA) | Token level (LCS points) | Egyptian, MSA | Their own manually annotated weblog data (context sensitive annotation) | F1-score=76.5% |
| Zaidan and Callison-Burch [11] | Language modeling | Word 3-gram LM | Sentence level | Egyptian, Gulf, Levantine, MSA (all dialects vs. MSA) binary classification | AOC dataset | Accuracy=77.8% |
| Zaidan and Callison-Burch [6] | Language modeling | Word 1-gram LM | Sentence level | Egyptian, Gulf, Levantine, MSA (all dialects vs. MSA) binary classification | AOC dataset | Accuracy=85.7% |
| Zaidan and Callison-Burch [6] | Language modeling | Character 5-gram LM | Sentence level | Egyptian, Gulf, Levantine, MSA (all dialects vs. MSA) binary classification | AOC dataset | Accuracy=85.0% |
| Lippincott et al. [97] | Prediction by Partial Matching (PPM) | Character features with maximal order N=3 | Document level (tweet Profiles) | 21 Arabic country dialects | MADAR Twitter corpus | Macro F1-score=50.43% |
| Huang [56] | Self-training | Geographical and text-based (word unigram) features | Sentence level | Egyptian, Gulf, Levantine, MSA | AOC dataset, Their own manually annotated Facebook data | Accuracy=84.4%, 65.5% |
| Huang [56] | Co-training | Geographical and text-based (word unigram) features | Sentence level | Egyptian, Gulf, Levantine, MSA | AOC dataset, Their own manually annotated Facebook data | Accuracy=86.2%, 67.7% |

### 6.2 Feature Engineering Supervised Approaches

The supervised machine learning algorithms build a statistical model of a set of data containing the inputs and the desired outputs (i.e., training data) [98]. In conventional supervised machine learning algorithms, the performance of models can be improved by extracting features from the raw data based on the domain knowledge. This process is called “feature engineering”. The increased availability of dialectal Arabic lexical resources as illustrated in Section 4 have resulted in enriching the conventional machine learning algorithms with feature engineering. The feature engineering supervised approaches frequently performed better than unsupervised and semi-supervised methods [6, 56, 44, 59]. The extracted features represent many aspects in the processed texts that aid in building high quality learning models. A considerable number of features have been investigated including surface features (word n-grams, character n-grams, a combination of word and character n-grams, word k-skip n-grams), grammatical features, dictionary-based features, meta features, and stylistic features. According to the research on feature engineering supervised approaches, word and character features' eligibility for the ADI task in the texts was extensively evaluated. Most of the variations between Arabic dialects are based on vocabularies and affixation. Therefore, most of the studies have investigated the use of character and word n-grams to capture lexical, sub-lexical (e.g., morphemes, affixes), and syntactic differences. The word-level and character-level n-gram features have proved in many studies to be effective for the task of ADI. Next, we will explain in detail the conventional machine learning algorithms formulated to solve the task of ADI in written texts, as well as the features utilized for training these algorithms.

#### 6.2.1 Naive Bayes

Cotterell and Callison-Burch [53] utilized word-level unigram, bigram, and trigram features along with two machine learning algorithms: Naive Bayes and linear SVM to train ADI models. The word unigram model outperformed the higher order models (bigrams and trigrams). The researchers linked the unigram model's success to the typical nonsequential order of dialectal words and the informal nature of dialectal texts. Furthermore, Naive Bayes outperformed a linear SVM, achieving an average accuracy of 87% in the pairwise classification of six Arabic varieties when tested on the Extended AOC dataset. In the study of Elaraby and Abdul-Mageed [59], similar to Cotterell and Callison-Burch's [53] findings, Naive Bayes outperformed other classical machine learning algorithms. That is, Naive bayes outperformed logistic regression and SVM with a linear kernel in the binary classification task (MSA vs dialects) and in the 3-way classification task (Egyptian vs. Levantine vs. Gulf). The sole exception was a linear SVM'a outperformance of Naive Bayes in the 4-way classification task (Egyptian vs. Levantine vs. Gulf vs MSA). Their experiments utilized a combination of word unigram, bigram and trigram features and were conducted under two different text representations, binary presence and term frequency inverse document frequency (TF-IDF). Most accurately, Naive Bayes yielded 84.53% (binary classification) and 87.81% (3-way classification), while SVM yielded 78.61% (4-way classification). Likewise, Elfardy and Diab [44] investigated Naive Bayes by building a model to identify sentences as either MSA or Egyptian. Their approach utilized token-level dialect labels from an underlying system for token-level identification of Egyptian dialectal Arabic developed by Elfardy et al. [5], in addition to perplexity-based features and meta features. Two word 5-grams language models for MSA and Egyptian Arabic were utilized to compute the perplexity of each sentence. The perplexity represents how confused the LM is about a given sentence while meta features estimate a sentence's degree of informality. Elfardy and Diab [44] study featured the total number of words, average word-length, percentage of words having word lengthening effects, and percentage of number, punctuation, special-characters and words written in Roman script. They also used binary features (yes/no) of whether 1) a sentence has consecutive repeated punctuation or not, 2) a sentence has an exclamation mark or not, and 3) a sentence has emoticons or not. The use of meta features with perplexity-based features and token-level ADI labels produced a highly accurate model. In addition, Elfardy and Diab conducted their experiments under various settings to examine the impact of two types of preprocessing techniques, tokenization and orthography normalization, as well as the impact of different sizes of the LM on their model's performance. Notably, tokenization helped address the data sparsity issue, producing better results than the surface experimental setting. Although orthographical conventions typically increase data sparseness and make it more difficult to estimate the true n-gram probabilities for language models, the orthography normalization utilized by Elfardy and Diab led to a slight improvement in the overall performance of the model. This can be attributed to the fact that orthography normalization may remove some dialectal cues. They also found that the performance of the model dropped when the size of the LM exceeded 16M tokens. They reported that as the size of the Egyptian and MSA LMs increased, the shared n-grams increased, resulting in higher confusability between the token classes in a given sentence. Their best model by far was based on Naive Bayes and trained using the aforementioned features after applying tokenization in the preprocessing step, having achieved 85.5% accuracy.

Sadat et al. [61] experimented to classify Arabic dialects into 18 dialect classes representing dialects spoken in 18 countries, as seen in Table 11. They experimented with Naive Bayes and character-level unigram, bigram, and trigram features. They implemented three n-gram models and found that character-based trigram and bigram features generally performed better than the character unigram features for most dialects. More specifically, the Naive Bayes models based on character bigram were more accurate than models based on character unigram and trigram. Their study also examined the Markov Language Model for Arabic dialect identification with cheater-level unigram, bigram, and trigram information. Their results showed that the character n-gram Naive Bayes outperformed the character n-gram Markov Language Model for most Arabic dialects. As best result, the Naive Bayes model based on character-based bigram features yielded an overall F1-score of 80% and 98% accuracy. Salameh et al. [99] was one of the first studies that explored Arabic fine-grained dialect identification at city level. Their research concerned 25 city-level Arabic dialects as well as MSA. They trained a Multinomial Naive Bayes (MNB) classifier with a combination of word unigram and character 1/2/3-gram features represented as TF-IDF. They also implemented word and character 5-gram language models for each dialect. The scores of these language models were used as additional features to train the classifiers. The classifier achieved 93.6% accuracy on MADAR Corpus-6 and 67.5% on MADAR Corpus-26. They also investigated the length of sentences and its correlation to accurately predicting the Arabic dialect in which it was written. They reported that their classifier managed to identify the exact city of a written text at an accuracy of 67.9% for sentences with an average length of 7 words and achieved around 90% for sentences with an average length of 16 words.

#### 6.2.2 Support Vector Machines (SVMs)

Many studies explored the use of SVM with various kernel functions and features for Arabic dialect identification. For example, the ASIREM system built by Adouane et al. [100] utilized the linear SVM with higher-order character-level n-grams where n $\in \{5,6\}$. Their system participated in the Discriminating between Similar Languages (DSL) shared task 2016 subtask 2, which handles Arabic dialect identification. The ASIREM system was ranked fourth in the closed track with an F1-score of 49.5% and was ranked first in the open track with 52.7% F1-score. The closed track allowed only the use of the corpora provided by the organizers while the open track allowed for additional data to be used. Most researchers who participated in the 2016 DSL subtask2 investigated the use of SVM with character-based n-grams [101, 102, 103, 104]. For instance the study of Ciobanu et al. [101] was ranked 8th out of 18 participants with an F1-score of 47.4%. They utilized an SVM with string kernels and character-level (2-7)-grams. Çöltekin and Rama [102] developed an SVM system using character (1-7)-gram features and participated in the Discriminating between Similar Languages (DSL) shared task 2016 subtask 2 of ADI in speech transcripts. Their system achieved an F1-score of 47.3% and was ranked 9th among 18 participants. Furthermore, Adouane et al. [89] utilized three methods: Cavnar's Text classification, linear SVM, and Prediction by Partial Matching (PPM). Adouane et al. [89] considered the automatic classification of Arabicized Berber (i.e., Berber written in Arabic script) as well as 7 Arabic dialects: Algerian, Egyptian, Gulf, Levantine, Mesopotamian, Moroccan, and Tunisian. Throughout the experiment they considered that three of these 7 dialects (i.e., Algerian, Moroccan, and Tunisian) are Arabic dialects at the country level, and the remaining dialects (i.e., Levantine, Mesopotamian, and Gulf) actually represent the dialects spoken in four regions: Egypt (Egyptian, Libyan, Sudanese), Levant (Jordanian, Lebanese, Palestinian, Syrian), Mesopotamia (Emirate, Iraqi, Kuwaiti, and Qatari), and Gulf (Bahrain, Oman, and Saudi Arabic). Plainly, slightly different divisions were made to count each North African variety as a stand-alone dialect and cluster Gulf/Mesopotamian dialect groups. The binary classification for each dialect was taken into account. The SVM outperformed the Cavnar's and the PPM methods, achieving an average F1-score of 92.94%. They constructed the SVM classifier with character-level 5-grams and 6-grams, as well as dictionary-based features. The lexicon of dialectal words was compiled from online resources, namely blogs, forums, and Facebook, using a script. Including the words of the compiled lexicon as a feature resulted in the best performance. Lastly, Adouane et al. [89] examined the problem of ADI in written texts at a document level. Their dataset has been compiled from online newspapers dedicated for dialectal Arabic, discussion forums, blogs, and Facebook. They have not divided the collected data into sentences, but considered each user comment/participation, and each paragraph in the online newspapers as a document.

#### 6.2.3 Decision Trees

Darwish et al. [92] used a Random Forest (RF) ensemble classifier that generates many decision trees, each of which is trained on a subset of features. They created an ensemble classifier that uses word unigram, bigram, and trigram models, as well as character unigram to 5-gram models as features. Their classifier achieved 83.3% accuracy. For training data, they used the Egyptian portion of the LDC2012T09 corpus [10]. For MSA training data, they utilized the MSAortion of the English/Arabic parallel corpus from the International Workshop on Arabic Language Translation[6], consisting of 150K sentences. Additionally, they manually created a test set by collecting and annotating 700 tweets (350 Egyptian tweets, and 350 MSA tweets). They also experimented with two more Random Forest (RF) ensemble classifiers that possess different features. The first classifier $S_{rule}$ trained on word and character n-gram features calculated from the segmented Egyptian training data. The segmentation was conducted based on manually predefined morphological rules. The $S_{rule}$ classifier reached 85.9% accuracy. The second classifier $S_{lex}$ relied on dialectal Egyptian lexicon-based features and their frequencies. They constructed three word lists: (1) 1,300 manually reviewed Egyptian words, (2) 94k manually reviewed Egyptian verbs, and (3) 8k manually reviewed Egyptian words with letter substitutions. The $S_{lex}$ classifier achieved the highest accuracy of 94.4%. Hence, Darwish et al. [92] concluded that the clean list of dialectal words that cover common dialectal phenomena is more efficient than the use of word and character n-grams. They also examined the use of character-level and word-level n-grams separately and reported that character-based n-gram features outperformed the word-based n-grams, because they generalized better to the new unseen test data where the lexical overlap between training and test data was low. However, the combination of character and word features resulted in better results than each one of them alone.

#### 6.2.4 Logistic Regression

Eldesouki et al. [105] experimented with several machine learning methods to perform a five-way classification among the five Arabic varieties, namely Egyptian, Gulf, Levantine, North-African, and MSA. They trained a one-vs-rest logistic regression model on character (1-5)-grams represented as TF-IDF. They evaluated their model on a development set and the model attained 66.19% accuracy. They also tried using word-level 1/2/3-gram features represented as TF-IDF, achieving 50.82% accuracy. The training and development data is made up of speech transcriptions derived from a multidialectal speech corpus created from high-quality broadcasts. The data has been used in the task of ADI in the 2017 VarDial Evaluation Campaign for Similar Languages, Varieties and Dialects. Elaraby and Abdul-Mageed [59] built an ADI classifier based on logistic regression to identify the dialects of four Arabic varieties, namely Egyptian, Levantine, Gulf, and MSA. They experimented with word 1/2/3-grams with two settings for feature representation: (a) presence vs. absence (1 vs. 0) vectors, and (b) TF-IDF vectors. In binary classification (dialectal Arabic vs. MSA), the model scored 83.71% and 83.24% accuracy for (1 vs. 0) and TF-IDF features representations respectively. In four-way classification (Egyptian vs. Levantine vs. Gulf vs. MSA), the model achieved the same results with 78.24% accuracy for the two different representations of features. They conducted their experiments using AOC dataset.

#### 6.2.5 Other Methods

String kernel functions have been used in text classification to measure the pairwise similarity between text samples, simply based on character n-grams [106]. String kernels along with kernel-based learning algorithms such as Kernel Discriminant Analysis (KDA) and Kernel Ridge Regression (KRR) have also been investigated to identify Arabic dialects and proved to be effective in many studies [107, 108, 109]. In fact, the system, based on multiple string kernels, was submitted to the Discriminating between Similar Languages (DSL) shared task in VarDial'2016 workshop and earned second place, achieving an accuracy of 50.91% and an F1-score of 51.31% [107]. Further contributing to this research, Ionescu and Butnaru [110] applied two transductive learning approaches to enhance the performance of string kernels. The first approach is based on interpreting the pairwise string kernel similarities between samples in both the training set and the test set as features. The second approach is a self-training technique of semi-supervised learning where a classifier is trained on the training set and tested on the test set. Then, a number of test samples with higher confidence scores are added to the training set for another round of training.

#### 6.2.6 Ensemble Methods

Dinu et al. [73] utilized an ensemble-based system to discriminate between dialects of Arabic using the corpus made available by the organizers of the third edition of ADI at the VarDial Evaluation Campaign 2018. They used a linear SVM for the individual classifiers and employed the majority rule to combine the output of the SVM classifiers. The individual classifiers were assigned uniform weights. They experimented with a set of features represented as TF-IDF: character n-grams, where $n \in \{1,...,8\}$, word n-grams where $n \in \{1,2,3\}$, and word k-skip bigrams where $k \in \{1,2,3\}$. They found that the optimal feature combination was character n-grams where n={3,4,5}. The best performing ensemble system yielded 0.5 F1-score on the test set. Their system did not participate in the third edition of the ADI shared task, but they used the same evaluation corpus and compared their system's performance with other systems submitted to the shared task. Their system's performance outperformed the task's baseline, but did not outperform other participating systems that were earned the first five places in the competition. Ragab et al. [111] developed an ensemble model of a group of best performing classifiers on a set of features that involves character-level and word-level TF-IDF features, class probabilities of a number of linear classifiers, and language model probabilities. To this end, they built 32 word 5-gram LMs of which 26 were trained on a corpus of 25 fine-grained city-level Arabic dialects, in addition to MSA. The remaining 6 LMs were trained on a corpus including MSA and only five Arabic dialects spoken in five cities across the Arab world. Likewise, Ragab et al. accomplished the same for 5-gram character level features by building 32 character 5-gram LMs. In addition to these 64 5-gram LMs, Ragab et al. [111] built 32 more LMs for Arabic varieties based on bi-gram word-level features. Each sentence in training and test dataset was scored by these 96 language models; then, the scores were scaled to 0-1. They used the scaled scores as input features to train a set of classifiers in two layers. The class (i.e., dialect) probabilities that resulted from the trained classifiers in the first layer were added to other features and used as input to train the second layer of classifiers. The classifiers in the second layer were then ensembled together by way of majority voting, which selects the most frequently detected dialect to be the final predicted dialect. The individual classifiers were built using the Multinomial Naive Bayes (MNB) technique. Their ensemble model achieved an F1-score of 67.20% and was ranked 3rd among 19 participating systems submitted to the MADAR shared task on fine-grained Arabic dialect identification 2019 task 1 about travel domain dialect identification. Malmasi et al. [57] conducted a set of experiments with a stacked generalization classifiers. The stacked generalization, or"stacking," involves creating an ensemble of individual classifiers by building a single linear SVM classifier for each feature type and utilizing the class probability outputs from each of these classifiers to build a higher level classifier, or "meta-classifier". The meta-classifier attempts to learn how to best combine the input predictions to produce a superior final output prediction. Their experiments were carried out to distinguish six Arabic dialects from each other, namely Egyptian, Jordanian, MSA, Palestinian, Syrian, and Tunisian, by performing 6-way Arabic dialect classification with various features. The Multidialectal Parallel Corpus of Arabic (MPCA) was used for training and testing. They evaluated character-level n-grams where n $\in$ {1,2,3,4} and word-level unigrams and bigrams with a single linear SVM. The features were weighted using TF-IDF. The character n-grams proved to be the best feature type, with trigrams yielding the highest accuracy of 65.26%. They also examined combining different feature types into a single feature vector. They reported that a combination of character 1/2/3-grams yielded the best results with an accuracy of 66.48%. They also tested the stacked generalization model with all feature types character and word combinations, achieving an accuracy of 74.32%. They assessed the generalization of the system and their learned features through a cross-corpus evaluation using three corpora: the MPCA, the AOC dataset, and the manually annotated 700 tweets by Darwish et al. [92]. They first trained a single model on the AOC and tested it on the MPCA dataset, and vice versa. They found that the character-level 1/2/3-gram features performed very well in both cross-corpus scenarios, achieving an accuracy of 83.35%, and 73.60% respectively. Then, they trained another model using both the MPCA and AOC datasets and tested it against Twitter data. They found that word unigrams yielded the best results with an accuracy of 96.71%. They suggested that character-level features generalize the most, but the word unigrams obtained the best performance with a large enough dataset. Hanani et al. [112] participated in the 2017 VarDial Evaluation Campaign on Natural Language Processing (NLP) for Similar Languages, Varieties and Dialects. They approached the shared task of ADI by combining multiple classifiers based on various machine learning algorithms. They submitted three runs. One of them combined four classifiers on the system level: SVM with a Radial Basis Function (RBF) kernel, Naive Bayes with multinomial distribution, logistic regression, and Random Forests with 300 trees. They achieved an F1-score equal to 31%. All these classifiers utilized the same features: character 1/2/3-grams presented to the system as a feature vector. Their best run, which achieved an F1-score of 62.8%, combined the text and acoustic features. That is, they used a focal multiclass model to combine the outputs of a word-based SVM multiclass model and an i-vector-based SVM multiclass model.

[6] `https://wit3.fbk.eu/mt.php?release=2013-01`

Generally speaking, traditional supervised learning approaches are effective in addressing the problem of ADI in written texts. The most widely used ML methods in many studies have been SVM, and Naive Bayes. However, the best results in the literature were obtained by SVM. Indeed, the SVM method with adequate features is suitable for the task of ADI in written texts, which is characterized by a large number of features and requires multi-class classification [13, 49, 112, 105, 69, 113, 102, 114, 115]. The ensemble methods also demonstrated good results for ADI in written texts, providing significant increase in performance for multi-class classification [13, 113]. Table 11 lists the Arabic Dialect Identification (ADI) models developed based on classical supervised approaches that rely on feature engineering.

### 6.3 Deep Supervised Approaches

Deep learning (DL) is a class of machine learning methods that employs multiple layers to progressively extract higher-level features from the raw input data. Deep learning has been applied to many fields including Natural Language Processing (NLP) and has achieved great success [116, 93]. The effectiveness of deep learning techniques in many NLP problems can be attributed partially to word embeddings, a distributional representation of texts that allows words with similar meaning to receive similar representations. In word embedding techniques, each word is represented as a real-valued vector with only tens or hundreds of dimensions, contrasting the millions of dimensions necessary for sparse word representation [93]. Researchers have sought out solutions in deep learning architectures, using word embedding algorithms such as GLoVe, FastText, and Word2Vec (Continuous Bag-of-Words (CBOW) model/Continous skip-gram model), in hopes of approaching the problem of ADI in written texts. The CBOW model can generate a word given its context, while the skip-gram model is used to generate context given a word [117]. Word sequences, character sequences, and combination of the two have been input into various deep learning architectures to perform Arabic dialect identification in written texts [118, 119, 120, 121, 122]. The DL has succeeded with NLP applications such as sentiment analysis and question classification [123, 124, 125, 126]. However, the findings of some studies demonstrated a poor performance of DL models; for example, the output of participating systems in Language Variety Identification of English, Spanish, and Portuguese in the 5th Author Profiling Task at PAN 2017 [122] demonstrated that systems based on traditional machine learning algorithms [127, 128, 129] outperformed those that relied on deep learning methods [130, 131, 132, 133]. Similarly, the deep-learning based methods used by most of the previously mentioned studies for the Arabic dialect identification task have performed poorly compared to the classical machine learning techniques with adequate feature engineering [102, 118, 134, 111, 135, 136]. In all of the four shared tasks organized for Arabic dialect identification from 2016 to 2019, the best performing systems were those that relied on classical machine learning algorithms and ensemble approaches with feature engineering. Meanwhile, the DL classifiers achieved lower results than many classical ML classifiers [69, 71, 72, 15]. This can be partially attributed to the limited resources used for Arabic dialect identification in terms of size, scope, and scale in these shared tasks; whereas DL models require a tremendous amount of annotated resources to adjust all parameters and reach high performance. Moreover, the MADAR shared task on fine-grained Arabic dialect identification handled Arabic dialects at city level with only 2,000 parallel sentences of 25 city-level Arabic dialects, which created a challenge for DL models, especially those that demand a considerable number of parameters to be set. Next, we explain the DL models implemented in the literature for Arabic dialect identification: their architectures and performances on various resources.

Zirikly et al. [137] experimented with neural networks trained with a 500-neuron single hidden layer and an output layer with softmax activation function. They used binary character (2-6)-gram representations as input features for the network. They also developed an ensemble classifier based on majority voting that took the outputs of one logistic regression and two single-layer Neural Networks and produced the majority label (i.e., Arabic dialects). The ties break when the output of the best-performing individual classifier is considered. They used character (1-6)-grams for LR and character (2-6)-grams for one of the NN. The second NN was trained using character (3-5)-grams and word unigram. Both of their single-layer NNs and the ensemble classifier participated in the Discriminating between Similar Languages (DSL) shared task 2016 subtask 2 of ADI in speech transcripts. They achieved F1-scores equal to 49.17%, and 49.22% respectively and the ensemble system was ranked 5th in the ADI subtask. Guggilla [138] also presented a system in the ADI subtask of the DSL 2016 based on Convolutional Neural Network (CNN) and was ranked 13th with an F1-score of 43.29%. They used a variant of the CNN architecture with four layers: the input, convolution, max-pooling, and softmax. Each sentence in the training data is represented in the input layer as a sentence comprised of distributional word embeddings. They initialized and utilized the randomly generated embeddings in the range $[-0.25 - 0.25]$. During training, they updated the input embedding vectors. More sophisticated neural networks and those that can be used on the front-end instead of word embeddings proved to be more useful for accurate classification of various Arabic dialects. Belinkov and Glass [119] developed a character-level CNN for ADI to be applied on the front-end to embed the sequence of characters into vector space. The sequence was then run through multiple convolutions, similar to a character-CNN utilized in language modeling [139]. The convolutional representations of texts were finally pooled to obtain a hidden vector representation of the text utilized to predict the Arabic dialect. Their system participated in the Discriminating between Similar Languages (DSL) shared task 2016, subtask 2. Their system achieved an F1-score of 48.34%, ranking 6th out of 18 participating systems. Ali [120] presented

Table 11: Summary of Arabic Dialect Identification (ADI) Models based on Feature Engineering Supervised Methods

| Reference | Learning Method/Model | Features Used | Processing Level | Class Labels | Evaluation Corpus | Evaluation Results |
|---|---|---|---|---|---|---|
| Cotterell and Callison-Burch [53] | Naive Bayes | Word unigram features | Sentence level | Pairwise classification (X vs. MSA) where X ∈ {Egyptian, Levantine, Gulf, Iraqi, Maghrebi, MSA} | Extended AOC dataset, Twitter | Average Accuracy=87%, 80.8% |
| Cotterell and Callison-Burch [53] | Naive Bayes | Word bigram features | Sentence level | Pairwise classification (X vs. MSA) where X ∈ {Egyptian, Levantine, Gulf, Iraqi, Maghrebi, MSA} | Extended AOC dataset, Twitter | Average Accuracy=81%, 80.8% |
| Cotterell and Callison-Burch [53] | SVM with a linear kernel | Word unigram features | Sentence level | Pairwise classification (X vs. MSA) where X ∈ {Egyptian, Levantine, Gulf, Iraqi, Maghrebi, MSA} | Extended AOC dataset, Twitter | Average Accuracy=85%, 77.4% |
| Elaraby and Abdul-Mageed [59] | Naive Bayes | Word 1/2/3-grams (binary presence) | Sentence level | Egyptian, Levantine, Gulf, MSA (all dialects vs. MSA) binary classification | AOC dataset | Accuracy=84.53% |
| Elaraby and Abdul-Mageed [59] | Naive Bayes | Word 1/2/3-grams (binary presence) | Sentence level | Egyptian, Levantine, Gulf (Egyptian vs. Levantine vs. Gulf) 3-way classification | AOC dataset | Accuracy=87.81% |
| Elaraby and Abdul-Mageed [59] | SVM | Word 1/2/3-grams (TF-IDF) | Sentence level | Egyptian, Levantine, Gulf, MSA (Egyptian vs. Levantine vs. Gulf vs. MSA) 4-way classification | AOC dataset | Accuracy=78.61% |
| Elfardy and Diab [44] | Naive Bayes | Token-level dialect labels, perplexity-based features, meta features | Sentence level | Egyptian, MSA | AOC dataset | Accuracy=85.5% |
| Sadat et al. [61] | Naive Bayes | Character-level bigram features | Sentence level | Algerian, Egyptian, Bahraini, Emirati, Iraqi, Jordanian, Kuwaiti, Lebanese, Libyan, Mauritanian, Moroccan, Omani, Palestinian, Qatari, Saudi, Sudani, Syrian, Tunisian | Their own manually annotated data from blogs and forums | F1-score=80% |
| Salameh et al. [99] | Multinomial Naive Bayes | Character 1/2/3-grams, word unigram, character/word 5-gram LM | Sentence level | 25 city-level Arabic dialects | MADAR travel domain corpus (Corpus-26) | Accuracy=67.5% |
| Salameh et al. [99] | Multinomial Naive Bayes | Character 1/2/3-grams, word unigram, character/word 5-gram LM | Sentence level | 25 city-level Arabic dialects | MADAR travel domain corpus (Corpus-6) | Accuracy=93.6% |
| Darwish et al. [92] | Random Forest (RF) ensemble classifier | Word 1/2/3-grams models, character (1-5)-gram models | Sentence level | Egyptian, MSA | Their own manually annotated 700 tweets | Accuracy=83.3% |

Table 11: Continued

| Reference | Learning Method/Model | Features Used | Processing Level | Class Labels | Evaluation Corpus | Evaluation Results |
|---|---|---|---|---|---|---|
| Darwish et al. [92] | Random Forest (RF) ensemble classifier | Word and character n-gram features calculated using manually created morphological rules | Sentence level | Egyptian, MSA | Their own manually annotated 700 tweets | Accuracy=85.9% |
| Darwish et al. [92] | Random Forest (RF) ensemble classifier | Egyptian lexicon-based features and their frequencies (three constructed word lists) | Sentence level | Egyptian, MSA | Their own manually annotated 700 tweets | Accuracy=94.4% |
| Adouane et al. [100] | linear SVM | Character 5+6 grams | Sentence level | Egyptian, Gulf, Levantine, North African, MSA | VarDial'2016 ADI corpus | Macro F1-score=49.5% (closed track), Macro F1-score=52.7% (open track) |
| Adouane et al. [89] | linear SVM | Character 5+6 grams and dictionary-based features | Document level | (X vs. remaining dialects) where X ∈ {Arabicized Berber, Algerian, Egyptian, Gulf, Levantine, Mesopotamian, Moroccan, Tunisian, MSA} | Their own manually annotated weblogs and online newspapers | Average F1-score=92.94% |
| Eldesouki et al. [105] | Logistic Regression | Character (1-5) grams (TF-IDF) | Sentence level | Egyptian, Gulf, Levantine, North African, MSA | VarDial'2016 ADI training data (80/20 train/development) | Accuracy=66.19% |
| Eldesouki et al. [105] | Logistic Regression | Word 1/2/3-grams (TF-IDF) | Sentence level | Egyptian, Gulf, Levantine, North African, MSA | VarDial'2016 ADI training data (80/20 train/development) | Accuracy=50.82% |
| Elaraby and Abdul-Mageed [59] | Logistic Regression | Word 1/2/3-grams (binary presence) | Sentence level | dialectal Arabic vs. MSA | AOC dataset | Accuracy=83.71% |
| Elaraby and Abdul-Mageed [59] | Logistic Regression | Word 1/2/3-grams (TF-IDF) | Sentence level | dialectal Arabic vs. MSA | AOC dataset | Accuracy=83.24% |
| Elaraby and Abdul-Mageed [59] | Logistic Regression | Word 1/2/3-grams (binary presence) | Sentence level | Egyptian, Levantine, Gulf, and MSA | AOC dataset | Accuracy=78.24% |
| Elaraby and Abdul-Mageed [59] | Logistic Regression | Word 1/2/3-grams (TF-IDF) | Sentence level | Egyptian, Levantine, Gulf, and MSA | AOC dataset | Accuracy=78.24% |
| Ionescu and Popescu [107] | Multiple string kernels | Character n-grams | Sentence level | Egyptian, Levantine, Gulf, North African, MSA | VarDial'2016 ADI corpus | Macro F1-score=51.31% |

Table 11: Continued

| Reference | Learning Method/Model | Features Used | Processing Level | Class Labels | Evaluation Corpus | Evaluation Results |
|---|---|---|---|---|---|---|
| Dinu et al. [73] | SVM ensemble system | Character (1-5)-grams | Sentence level | Egyptian, Levantine, Gulf, North African, MSA | VarDial'2018 ADI corpus | F1-score=50% |
| Ragab et al. [111] | Ensemble model of a group of best performing classifiers on a set of various features | Character-level and word-level TF-IDF 5-grams, class probabilities of a number of linear classifiers, and language model probabilities | Sentence level | 25 city-level Arabic dialects, MSA | MADAR travel domain corpus (Corpus-26) | Macro F1-score=67.20% |
| Malmasi et al. [57] | linear SVM | Character trigrams (TF-IDF) | Sentence level | Egyptian, Syrian, Jordanian, Palestinian, Tunisian, MSA | MPCA dataset | Accuracy=65.26% |
| Malmasi et al. [57] | linear SVM | Character (1-3)-grams (TF-IDF) | Sentence level | Egyptian, Syrian, Jordanian, Palestinian, Tunisian, MSA | MPCA dataset | Accuracy=66.48% |
| Malmasi et al. [57] | Stacked generalization model trained on MPCA dataset | Character (1-4)-grams, word unigrams and bigrams (TF-IDF) | Sentence level | Egyptian, Syrian, Jordanian, Palestinian, Tunisian, MSA | MPCA dataset | Accuracy=74.32% |
| Malmasi et al. [57] | Stacked generalization model trained on MPCA dataset | Character (1-3)-grams (TF-IDF) | Sentence level | Egyptian, Syrian, Jordanian, Palestinian, Tunisian, MSA | AOC dataset | Accuracy=73.60% |
| Malmasi et al. [57] | Stacked generalization model trained on AOC dataset | Character (1-3)-grams (TF-IDF) | Sentence level | Egyptian, Syrian, Jordanian, Palestinian, Tunisian, MSA | MPCA dataset | Accuracy=83.35% |
| Malmasi et al. [57] | Stacked generalization model trained on AOC + MPCA dataset | Word unigrams (TF-IDF) | Sentence level | Egyptian, Syrian, Jordanian, Palestinian, Tunisian, MSA | Twitter dataset created by [92] | Accuracy=96.71% |
| Hanani et al. [112] | Multiple combined classifiers: SVM with a Radial Basis Function (RBF) kernel, Naive Bayes with multinomial distribution, logistic regression, and Random Forests with 300 trees | Character 1/2/3-grams | Sentence level | Egyptian, Levantine, Gulf, North African, MSA | VarDial'2017 ADI corpus | Weighted F1-score=31% |
| Hanani et al. [112] | Focal multiclass model combining the outputs of a word-based SVM multiclass model, and an i-vector-based SVM multiclass model | Text and acoustic features | Sentence level | Egyptian, Levantine, Gulf, North African, MSA | VarDial'2017 ADI corpus | Weighted F1-score=62.8% |

their system to the Arabic Dialect Identification (ADI) shared task at the VarDial Evaluation Campaign 2018. They proposed the use of character-level CNN, a neural network classifier that uses both the transcript text as one-hot encoded sequence of characters (padded or truncated to match a predefined maximum length) and the corresponding dialect embedding feature vector. The character sequence passes through a series of five layers (Gated Recurrent Units (GRU), convolution, batch-normalization, max-pooling, and dropout) before finally reaching a softmax layer. Then, the embedding vector passes directly to the softmax layer. The outputs of both softmax layers are averaged to provide the final output, which represents the probability distribution over five Arabic dialects. Their proposed character-level CNN achieved an F1-score equal to 57.6% and was ranked the 2nd among 6 participants.

Fares et al. [140] introduced two ensemble systems constructed from multiple neural networks based models as well as the Multinomial Naive Bayes classifier. The first ensemble system consists of three individual models. The first and second models are DL-based models built to combine both character and word features. The first model takes one-hot encoded sequence of characters to Long Short-Term Memory (LSTM) (for capturing context for character features), then through five convolution layers, and finally to a softmax layer (for calculating probabilities). The second model consists of an embedding layer of fastText word embeddings [141], an LSTM (for capturing context for word features), and a softmax layer. The output of the softmax layers of both DL models are averaged to produce the final output. The third model is a Multinomial Naive Bayes (MNB) with character (1-5)-gram and word unigram TF-IDF features. The ensemble system achieved F1-scores equal to 65.35%. The second ensemble system consists of the MNB classifier and a DL model. The DL model has two hidden fully connected layers and an output layer. The DL model takes as input the frequency-based features of MNB classifier's features (character (1-5)-gram and word unigram TF-IDF features). The second ensemble system achieved an F1-scores equal to 65.66%. They participated in the MADAR Shared Task on Arabic Fine-Grained Dialect Identification task 1 on travel domain. Their second ensemble system was ranked 7th among 19 participating systems. Elaraby and Zahran [142] participated in the MADAR shared task on Arabic fine-grained dialect identification task 2 regarding dialect identification for Twitter users. Their system was trained using Convolutional Bidirectional LSTM (C-BiLSTM. Their DL model consists of four layers: the input, convolution, BiLSTM, and Softmax. The input layer maps a sequence of words into a real-valued sequence vector while character embedding is randomly initialized and not learned externally. Their system achieved an F1-score of 61.5% and was ranked 4th among the 9 participating systems in the MADAR Twitter user dialect identification task. Their study as well as the study of Samih et al. [121] reported that training on concatenated user tweets as input to the DL model is superior to training on user tweets independently. The ensemble system of Lippincott et al. [97] merged several neural architectures: CNN, Recurrent Neural Networks (RNN), and Multi-Layer Perceptron (MLP). They used character and word sequences for CNN and RNN. The probability distributions from language models and metadata were used as input to MLP. The ensemble system concatenates the hidden representations produced by the DL models (i.e., CNN, RNN, and MLP) and stacks one or more layers, stepping down in size to a final softmax output layer. Their system participated in the MADAR Travel domain dialect identification task, achieving an F1-score of 61.83% and was ranked 13th out of the 19 participants. Francony et al. [14] presented a DL method for Arabic fine-grained dialect identification. They implemented a hierarchical model of two levels of DNNS in which the first predicts the group of the dialects (i.e., region) and the second level predicts the exact fine-grained dialect (i.e., city) based on the region prediction. The DNN in the first level consists of three layers: a B-LSTM, followed by a fully connected layer, and then an output layer. The second level is actually a set of 7 DNNs, one for each region. The input of the model is produced using Word2Vec. The F1-score of the hierarchical DL model on the MADAR travel domain corpus is equal to 58%. Table 12 summarizes the Arabic dialect identification models developed using deep learning approaches.

## 7 Open Issues and Future Directions

Arabic dialect identification started to gain a great deal of attention in the field of Arabic NLP a decade ago with the heightened prevalence of Web 2.0 and the rapid growth of online user-generated contents. Although a considerable number of studies have been conducted so far with beneficial results for Arabic dialect identification, most of the work has been devoted to the five main groups of Arabic dialects, namely Egyptian, Gulf, Levantine, North African, and MSA (see Table 13). The Arabic dialect taxonomy is complex and there are many overlapping areas between Arabic dialects within the same region and even the same country. Therefore, a lot of work remains to be carried out for fine-grained Arabic dialect identification, including both approaches and large-scale annotated resources, as there are few publicly available ADI annotated corpora. Moreover, the publicly available corpora are limited in terms of size, scope, and scale, which creates restricts deep empirical comparisons between approaches. Much larger-scale and larger-scope

Table 12: Summary of Arabic Dialect Identification (ADI) Models based on Deep Supervised Methods

| Reference | Architecture | Features Used | Processing Level | Class Labels | Evaluation Corpus | Evaluation Results |
|---|---|---|---|---|---|---|
| Zirikly et al. [137] | Single-layer Neural Network consisting of single hidden layer of 500 neurons and output layer with softmax activation function | Binary character (2-6)-grams representations | Sentence level | Egyptian, Levantine, Gulf, North African, MSA | VarDial'2016 ADI corpus | Macro F1-score=49.17% |
| Zirikly et al. [137] | Ensemble classifier based on majority voting that took the outputs of one logistic regression and two single-layer Neural Networks | Character (1-6)-grams for LR, character (2-6)-grams for 1st NN, character (3-5)-grams and word unigram for 2nd NN | Sentence level | Egyptian, Levantine, Gulf, North African, MSA | VarDial'2016 ADI corpus | Macro F1-score=49.22% |
| Guggilla [138] | A variant of the CNN architecture (an input layer, a convolution layer, a max pooling layer, and a fully connected softmax layer) | Randomly generated embeddings in the range $[-0.25, 0.25]$ and updated during training | Sentence level | Egyptian, Levantine, Gulf, North African, MSA | VarDial'2016 ADI corpus | Macro F1-score=43.29% |
| Belinkov and Glass [119] | Character-level CNN (embedding layer, dropout layer, multiple parallel convolutional layers with different filter widths, max pooling layer, fully-connected layer, and a softmax layer) | Character embedding learned during training | Sentence level | Egyptian, Levantine, Gulf, North African, MSA | VarDial'2016 ADI corpus | Macro F1-score=48.34% |
| Ali [120] | Character-level CNN with two inputs (1) one-hot encoded sequence of characters going through a series of layers (GRU layer, convolutional layer, batch-normalization layer, max-pooling layer, dropout layer, and softmax layer), (2) dialect embedding feature vector going directly to the softmax layer | One-hot encoded sequence of characters and dialect embedding feature vector | Sentence level | Egyptian, Levantine, Gulf, North African, MSA | VarDial'2018 ADI corpus | Macro F1-score=57.6% |

Table 12: Continued

| Reference | Architecture | Features Used | Processing Level | Class Labels | Evaluation Corpus | Evaluation Results |
|---|---|---|---|---|---|---|
| Fares et al. [140] | Ensemble system: (1) LSTM + CharCNN, (2) FastText embeddings+LSTM, (3) MNB classifier | DL models: one-hot encoded sequence of characters and word embeddings; MNB classifier: (1-5)-grams character and unigram word features using TF-IDF | Sentence level | 25 city-level Arabic dialects, MSA | MADAR travel domain corpus (Corpus-26) | Macro F1-score=65.35% |
| Fares et al. [140] | Ensemble system: (1) (Character TF-IDF) + (Word TF-IDF) + NN (2) MNB classifier | MNB classifier: (1-5)-grams character and unigram word features using TF-IDF; NN: frequency-based features of MNB classifier's features | Sentence level | 25 city-level Arabic dialects, MSA | MADAR travel domain corpus (Corpus-26) | Macro F1-score=65.66% |
| Elaraby and Zahran [142] | C-BiLSTM (input layer, convolution layer, BiLSTM Layer, and Softmax Layer) | Word and character embeddings | Document level (tweet Profiles) | 21 Arabic country dialects | MADAR Twitter corpus | Macro F1-score=61.5% |
| Lippincott et al. [97] | Ensemble system of 3 DL models (CNN, RNN, and MLP) concatenates the hidden representations produced by the three DL models and stacks one or more layers, stepping down in size to a final softmax output layer. | Character and word embeddings, language-model based features | Sentence level | 25 city-level Arabic dialects, MSA | MADAR travel domain corpus (Corpus-26) | Macro F1-score=61.83% |
| Francony et al. [14] | Hierarchical system of two levels. The 1st level is a DNN with B-LSTM, followed by fully-connected layer, and then output layer. The 2nd level is a set of 7 different DNNs each of which utilizes a RNN layer, followed by fully-connected layer, and then output layer | Word embeddings | Sentence level | 25 city-level Arabic dialects, MSA | MADAR travel domain corpus (Corpus-26) | Macro F1-score=58% |

Table 13: The Arabic dialects covered in the ADI literature.

Coarse-grained ↓ Fine-grained

| | Arabic dialects | References |
|---|---|---|
| Region-based | Egyptian, MSA | Elfardy et al., 2013; Elfardy and Diab, 2013; Tillmann et al., 2014; Darwish et al., 2014; Elfardy et al., 2014; Al-Badrashiny et al., 2015 |
| Region-based | Egyptian, Gulf, Levantine, MSA | Zaidan and Callison-Burch, 2011; Tillmann et al., 2014; Elfardy and Diab, 2012; Al-Badrashiny and Diab, 2016; Elaraby and Abdul-Mageed, 2018; Lulu and Elnagar, 2018; Huang, 2015 |
| Region-based | Egyptian, Gulf, Levantine, North African, MSA | Malmasi and Zampieri, 2016; Zirikly et al., 2016; Dinu et al., 2018; Ionescu and Butnaru, 2018; Ionescu and Butnaru, 2017; Çöltekin and Rama, 2017; El-Haj et al., 2018; Zampieri et al., 2017; Michon et al., 2018; Malmasi et al., 2016; Ali, 2018; Zampieri et al., 2018 |
| Region-based | Egyptian, Gulf, Levantine, North African, Iraqi, MSA | Cotterell and Callison-Burch, 2014 |
| Country-based | Algerian, Egyptian, Gulf, Levantine, Mesopotamian, Moroccan, Tunisian | Adouane et al., 2016b |
| Country-based | Algerian, Moroccan, Palestinian, Syrian, Tunisian, MSA | Lichouri et al., 2018 |
| Country-based | Jordanian, Lebanese, Palestinian, Syrian | Wray, 2018 |
| Country-based | Egyptian, Jordanian, Palestinian, Syrian, Tunisian, MSA | Bouamor et al., 2014; Malmasi et al., 2015 |
| Country-based | Algerian, Egyptian, Bahraini, Emirati, Iraqi, Jordanian, Kuwaiti, Lebanese, Libyan, Mauritanian, Moroccan, Omani, Palestinian, Qatari, Saudi, Sudanese, Syrian, Tunisian | Sadat et al., 2014 |
| City-based | Annaba, Algiers, Sfax, Palestinian, Syrian | Harrat et al., 2015 |
| City-based | Tenes, Constantine, Djelfa, Ain-Defla, Tizi-Ouzou, Batna, Annaba, Algiers | Lichouri et al., 2018 |
| City-based | Rabat, Fes, Algiers, Tunis, Sfax, Tripoli, Benghazi, Cairo, Alexandria, Aswan, Khartoum, Jerusalem, Amman, Salt, Beirut, Damascus, Aleppo, Mosul, Baghdad, Basra, Doha, Mascat, Riyad, Jeddah, Sana'a | Salameh et al., 2018; Bouamor et al., 2018 |

annotated corpora are needed to enrich the ADI resources and to support qualitative comparisons between potential studies in this area. There is also a need to criticize the available resources and analyze them in order to find the gaps in the available ADI resources.

The main sources of dialectal Arabic in its written form are online blogs, discussion forums and social networks. These online contents are written informally with many spelling errors. They include a lot of speech effects, emojis, neologisms, elongations and are written using different scripts. Therefore, NLP tools that handle these issues and clean the texts are needed to fill the gaps and help advance the research in dialectal Arabic. In fact, more investigations are required to handle the dialectal Arabic challenges that affect dialectal Arabic processing in general, and Arabic dialect identification in particular. We noticed that few studies on ADI investigated the impact of properly pre-processing dialectal Arabic contents (e.g., tokenizations, orthography normalization, stemming) and cleaning them (e.g., removing stop words, correcting spelling errors) before building and training ADI models. More investigations are also necessary to handle ADI challenges themselves. For example, there have been only a handful studies examining the Linguistic Code Switching (LCS) phenomenon where a speaker mixes two or more Arabic varieties in the same utterance. Other challenges remain such as handling Arabizi, a non-standard romanization used by some Arabic native speakers for online contents, and the rich cliticization system in many Arabic dialects. Continued intensive investigation of these challenges will help uncover new appropriate approaches to distinguish between Arabic dialects.

## 8 Conclusion

The ADI in written texts plays fundamental role in many cross-language NLP applications, as well as social media analysis. It is considered the first step in building intelligent language systems that handle online contents. The task of ADI in written texts is a complex problem, as there are a considerable number of Arabic dialects based on various levels of geographical location (e.g., classification based on region, country, city). Many other factors actually affect the appearance of Arabic dialects, even within the same geographical location such as lifestyle, education, and socioeconomic status. The limited amount of training data available for Arabic dialects, especially at the fine-grained levels, as well as the necessity to deal with online dialectal Arabic contents have attracted many researchers of Arabic NLP in the last decade and more exponentially in the last four years to investigate the problems facing ADI in written texts.

This paper presented an extensive overview of ADI studies in literature. The algorithmic learning methods examined to perform ADI were discussed starting from conventional machine learning techniques used in early studies on ADI, up until neural networks and deep learning methods. We briefly compared various proposed methods. Our survey also discussed in detail the features used in ADI studies and their efficiency in improving the overall performance of the implemented systems, as well as the techniques used to represent these features. The available Arabic dialect identification corpora required for building ADI models were also covered in the survey, along with the common benchmarks used in the literature to evaluate the models. Future work and remaining open issues were discussed to help advance the field of dialectal Arabic NLP in general and ADI in written texts specifically.